\documentclass{article} 
\usepackage{iclr2027_conference,times}

\usepackage{amsmath,amsfonts,bm}

\def\eqref#1{equation~\ref{#1}}

\def\1{\bm{1}}

\DeclareMathAlphabet{\mathsfit}{\encodingdefault}{\sfdefault}{m}{sl}
\SetMathAlphabet{\mathsfit}{bold}{\encodingdefault}{\sfdefault}{bx}{n}

\iclrfinalcopy

\usepackage{hyperref}
\usepackage{url}
\usepackage[utf8]{inputenc} 
\usepackage[T1]{fontenc}    
\usepackage{booktabs}       
\usepackage{amsfonts}       
\usepackage{nicefrac}       
\usepackage{microtype}      
\usepackage{xcolor}         

\usepackage{graphicx}
\usepackage{multirow}
\usepackage{subcaption}
\usepackage{tcolorbox}
\usepackage{amsmath} 
\usepackage{makecell}
\usepackage{adjustbox}
\usepackage{xspace}
\usepackage{arydshln}
\usepackage{algorithm}
\usepackage{mathrsfs}
\usepackage{setspace}
\usepackage[noend]{algpseudocode}
\usepackage{amssymb}
\usepackage{mathtools}
\usepackage{amsthm}
\usepackage[inkscapelatex=false]{svg}
\usepackage{thmtools} %
\usepackage{caption} %
\usepackage{float}
\usepackage{wrapfig}

\title{Rethinking the Evaluation of Harness \\ Evolution  for Agents}

\author{%
    Yike Wang\thanks{equal contribution} \ $^{1,2}$ \
    Huaisheng Zhu\footnotemark[1] \ $^3$\
    Zhengyu Hu$^2$\
    Yige Yuan$^2$\\
   \textbf{Zhengyu Chen$^3$ \ Shakti Senthil$^2$ \ Hannaneh Hajishirzi$^2$ \ Yulia Tsvetkov$^2$ } \\
   \textbf{Noah A. Smith$^{1,2}$ \ Pradeep Dasigi$^1$ \ Teng Xiao\footnotemark[1] \ $^1$} \\
    $^1$Allen Institute for AI\;
    $^2$University of Washington\;
    $^3$Independent \ \\
    \texttt{yikewang@cs.washington.edu, \{zhuhs1998, tengxiao01\}@gmail.com}
}

\begin{document}

\maketitle
\lhead{\textit{preprint}}

\begin{abstract}
Harness evolution is an iterative search procedure that repeatedly evaluates and revises candidate harnesses used for LLM agents using task feedback. We revisit the evaluation of such automatic harness evolution procedures and identify two fundamental issues in the protocol. First, prior work does not compare these approaches with simple task-level search baselines under matched feedback and inference budgets. Second, prior work searches for harness configurations using verification signals (e.g., unit test cases) drawn from the same benchmarks on which it reports the final performance of the evolved harnesses, violating the standard separation between training and test data.
To address this, we compare automatic harness evolution with simple test-time scaling and discovery baselines under comparable feedback and inference budgets, and evaluate evolved harnesses on held-out tasks to assess generalization. Following prior work, we experiment on Terminal-Bench~2.1 and find that automatic harness evolution fails to outperform simple test-time scaling methods both with and without test cases, and exhibits limited generalization.
However, we find that long-horizon games are a promising setting for automatic harness evolution, as they are difficult enough to leave headroom, rely heavily on adaptation to out-of-distribution dynamics, and provide granular feedback by design. In these settings, task-specific harness evolution improves over the search baseline by 80\% on ARC-AGI-3 and by 11\% on EdgeBench games under matched budgets.  Together, these findings highlight the need for matched-budget baselines and held-out evaluation to distinguish genuine harness improvements from benchmark-specific search and overfitting, and point to a more careful characterization of when automatic harness evolution is actually useful. 
Our code is available at \url{https://github.com/rethinking-harness-evolution}.
\end{abstract}

\section{Introduction}
Large language model (LLM) agents increasingly rely on external harnesses to interact with complex environments~\citep{yang2024swe, merrill2026terminal}. A harness defines the prompts, tools, memory, verification routines, and control logic through which a model observes tasks and acts~\citep{lopopoloHarnessEngineeringLeveraging2026}. Prior work shows that harness engineering can substantially affect agent performance even with a fixed model~\citep{trivedyImprovingDeepAgents2026}, yet until recently it remained largely manual: developers inspect trajectories, diagnose failures, and revise the harness. This has motivated automatic harness evolution, in which agents improve external harnesses~\citep{leeMetaHarnessEndtoEndOptimization2026, lin2026agentic, zhang2026harnessing}. These methods typically follow a search loop: analyze prior trajectories, scores, and failures; propose modifications; evaluate them on benchmark tasks; and repeat~\citep{leeMetaHarnessEndtoEndOptimization2026, zhang2026harnessing}.

We identify two fundamental problems with current evaluation of harness evolution. 
First, existing studies~\citep{leeMetaHarnessEndtoEndOptimization2026,lin2026agentic,zhang2026harnessing} do not compare these approaches against simple task-level search baselines under matched feedback and inference budgets. Such baselines spend additional computation on the evaluation tasks themselves, for example through parallel sampling or sequential refinement~\citep{snell2024scaling,li2026benchmark,novikov2025alphaevolve}.
Second, these works search for harness configurations using verification signals (e.g., unit test cases) drawn from the same benchmarks on which they report the final performance of the evolved harnesses, violating the standard separation between training and test data.
The existing evaluation protocol therefore leaves a fundamental question unresolved: \textit{Does harness evolution yield generalizable improvements in harness design, or are its gains primarily due to repeated sampling?} 


To address these concerns, we compare parallel sampling, sequential refinement, and harness evolution under a controlled budget protocol that specifies what feedback each method receives, how it allocates inference compute, and whether it modifies task trajectories or the harness itself. Following prior work, we evaluate these methods on Terminal-Bench~2.1~\citep{merrill2026terminal}  
with multiple models. We find that automatic harness evolution fails to outperform simple test-time scaling baselines, both when verification signals are available and when they are not. Moreover, when the same set of tasks is used for both harness evolution and final evaluation, reported gains overstate true harness improvements, as they reflect adaptation to specific instances rather than generalizable harness design.

Further analysis reveals that long-horizon games are a promising setting for automatic harness evolution, as they are difficult enough to leave headroom, rely heavily on adaptation to out-of-distribution dynamics, and provide granular feedback by design. Starting from well-engineered harnesses, automatic harness evolution improves over fixed-harness sequential refinement by 80\% on ARC-AGI-3~\citep{arcprize2026arcagi3} and by 11\% on EdgeBench games~\citep{zhu2026edgebench} under matched budgets.
Together, these findings call for evaluation protocols that separate optimization feedback from final measurement and compare against test-time scaling baselines, and suggest that harness evolution may benefit adaptation to out-of-distribution, long-horizon tasks with rich feedback.

\begin{section}{Related Work}\label{sec:related}
\paragraph{Test-time Scaling}
Test-time scaling refers to improving a language model's outputs by allocating additional computation at inference, without modifying the model weights~\citep{snell2024scaling}. Existing work falls into two broad categories: parallel sampling~\citep{wang2022self, brown2024large} and sequential refinement~\citep{madaan2023self}. Parallel sampling draws multiple candidate answers independently and aggregates them by voting or verifier reranking. Sequential refinement generates answers iteratively, conditioning each attempt on prior ones to revise earlier reasoning. 
In this work, we argue that test-time scaling methods serve as strong and meaningful baselines for harness evolution algorithms.

\paragraph{Automatic Harness Evolution}
A growing body of work improves agent performance without retraining the underlying model by optimizing the external scaffold around the model. Early efforts target a single editable surface: prompt and instruction optimizers tune prompts, demonstrations, or language-model programs from execution feedback~\citep{khattab2023dspy, opsahl2024optimizing, yuksekgonul2025optimizing, agrawal2025gepa}, while in-context methods accumulate experience as natural language artifacts that the model can reuse at inference time~\citep{zhao2024expel, zhang2025agentic}. These approaches show that agent experience can be externalized outside model weights, but they typically optimize one component at a time. A more recent line treats the full harness as the object of search. Meta-Harness~\citep{leeMetaHarnessEndtoEndOptimization2026} searches over harness code using prior source code, scores, and execution traces. Agentic Harness Engineering~\citep{lin2026agentic} evolves prompts, tools, middleware, skills, and long-term memory through an observability-driven loop. AEVO~\citep{zhang2026harnessing} views agentic evolution itself as an interactive environment and edits the procedure or agent context that controls future search. Our work studies the evaluation of automatic harness evolution, comparing it against test-time scaling baselines under matched feedback and inference budgets, and testing whether evolved harnesses transfer to held-out tasks from the same benchmark.
\end{section}

\section{Rethinking the Evaluation of Harness Evolution}\label{sec:method}
In this section, we formalize the evaluation of harness evolution under a unified budget view. The goal is to make explicit whether performance gains reflect reusable harness improvement or test-time discovery over the evaluated tasks. Given a fixed agent policy, a task distribution, and a compute budget, each method is specified by what it updates and what feedback it observes. We consider four methods: parallel sampling, sequential refinement, harness evolution, and harness scaling. They spend the budget on independent trajectories, trajectory revisions, shared harness updates across tasks, and task-specific harness updates, respectively. We first introduce the common notation, and then describe each method in Sections~\ref{sec:method_parallel_sampling}--\ref{sec:method_harness_scaling}.

\begin{figure*}[t]
\centering
\includegraphics[width=1.0\linewidth]{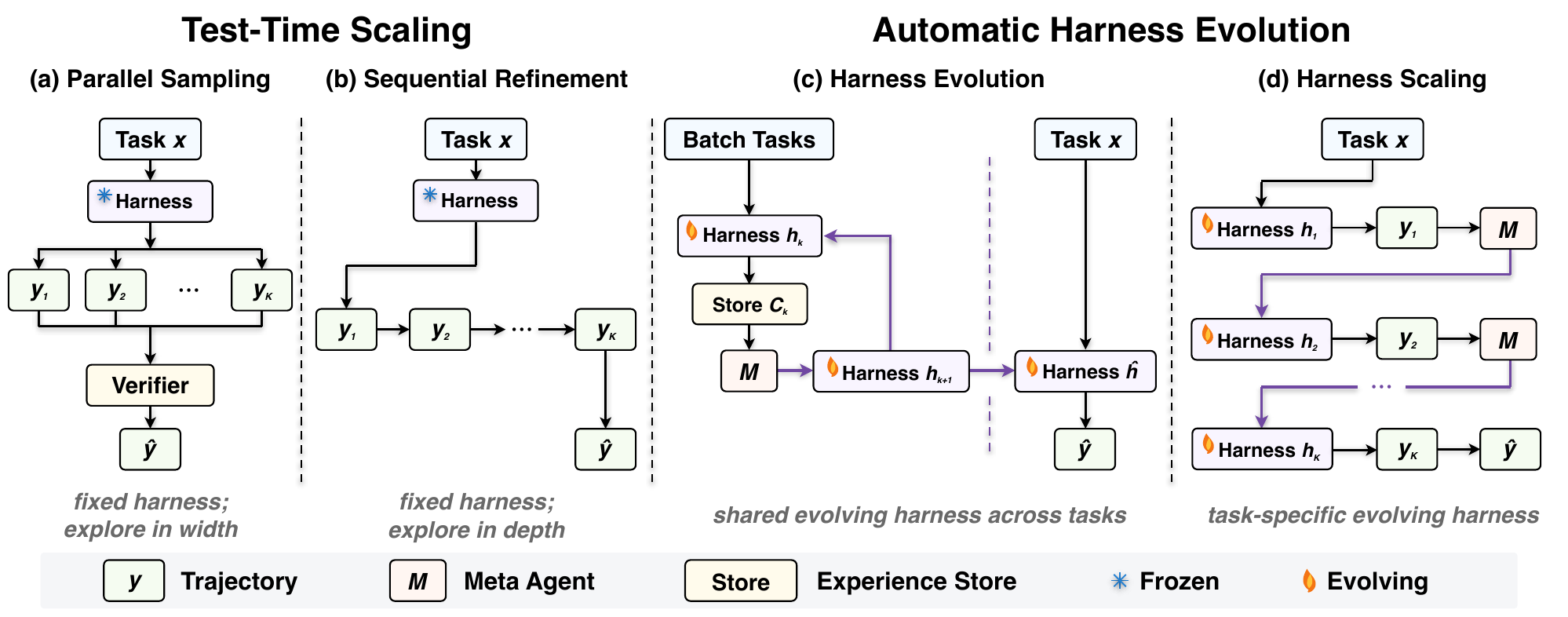}
\caption{\textbf{Test-time scaling and automatic harness evolution algorithms.} We compare four approaches under a unified budget, emphasizing what feedback the agent receives, how it spends its budget, and whether it modifies the trajectories or the harness.}
\vspace{-2mm}
\label{fig:framework}
\end{figure*}

\subsection{Preliminary}
Let $\pi_\theta$ denote the agent policy and let $x$ denote a task drawn from a task distribution $\mathcal{X}$. Executing the agent on $x$ using the harness $h$ yields a trajectory $y$, the full sequence of states, actions, and observations, which we write as $y \sim \pi_\theta(\cdot \mid x; h)$. When a unit test case $g$ is available, we define an outcome $R(y, g) \in \{0, 1\}$ that returns $1$ exactly when the trajectory solves the task. Every method below consumes a compute budget $K$ and returns a final trajectory $\hat{y}$ for each task $x \sim \mathcal{X}$. 

We also define a summarization map $\Phi$, realized by the same underlying model as the policy. The map $\Phi$ summarizes an experience store into a form that a downstream agent can consume, such as recurring failures, redundant attempts, or cost, so that subsequent refinement or editing steps can diagnose why earlier attempts underperformed and avoid repeating edits. 

\subsection{Parallel Sampling}
\label{sec:method_parallel_sampling}
Parallel sampling (Figure~\ref{fig:framework} (a)) is a representative test-time scaling method that generates multiple trajectories in parallel~\citep{wang2022self, brown2024large}. Specifically, we draw $K$ trajectories independently from the policy using a fixed harness $h$,
\[
  y_1, \dots, y_K \sim \pi_\theta(\cdot \mid x; h),
\]
such that exploration is spread across $K$ mutually independent attempts. Increasing $K$ expands the reachable action space, thereby improving the likelihood that the agent explores at least one trajectory that contains a correct solution. We consider two settings. First, when unit test cases are not available, selection instead relies on model self-selection, where the model is asked to choose the best candidate among the sampled solutions. Let $J$ be a self judge realized by the agent itself, and we return the highest scoring candidate,
\[
  \hat{y} = \operatorname*{arg\,max}_{k \le K} J(y_k).
\]
Second, we assume access to unit test cases, which provide verifiable signals that can be used directly to select the final response. In this case, we set $\hat{y}$ to any trajectory accepted by the outcome,
\[
  \hat{y} \in \{\, y_k : R(y_k, g) = 1 \,\}.
\]

\subsection{Sequential Refinement}
\label{sec:method_sequential_refinement}
In contrast to parallel sampling, sequential refinement~\citep{madaan2023self}  (Figure~\ref{fig:framework} (b)) allocates the budget to depth, extending the interaction horizon so that each trajectory builds on the last. In this process, the model leverages prior reasoning experience and explores alternative solution paths to guide refinement in subsequent turns. To realize this process, we sample the first trajectory directly based on the prompt,
\[
  y_1 \sim \pi_\theta(\cdot \mid x; h),
\]
and for $k = 2, \dots, K$, we condition the next trajectory on the previous one, which lets the agent reflect on and revise its prior attempt, using the same fixed harness $h$. When no unit test case is available, the agent reads the previous trajectory alone,
\[
  y_k \sim \pi_\theta\big(\cdot \mid x,\ \Phi(y_{k-1});\ h\big),
\]
and we return the final trajectory in the sequence, $\hat{y} = y_K$. 

When a unit test case is available, we expose both the trajectory and its outcome, so that the agent can refine against the signal for the next turn's response,
\[
  y_k \sim \pi_\theta\big(\cdot \mid x,\ \Phi(y_{k-1}, R(y_{k-1}, g));\ h\big).
\]
In this case, we return any trajectory accepted by the outcome,
\[
\hat{y} \in \{\, y_k : R(y_k, g) = 1 \,\}.
\]

\subsection{Harness Evolution}
\label{sec:method_harness_evolution}
Harness evolution methods~\citep{leeMetaHarnessEndtoEndOptimization2026, lin2026agentic, zhang2026harnessing}  (Figure~\ref{fig:framework} (c)) retain the iterative loop of sequential refinement but allocate budget to the harness surrounding the policy rather than to trajectories alone, and optimize across a task distribution rather than fitting to one task. We draw a batch of tasks $\{x^{(i)}\}_{i=1}^{n} \sim \mathcal{X}$ and initialize the experience store with them, $\mathcal{C}_0 = \{x^{(i)}\}_{i=1}^{n}$. At round $k$ we sample $m$ rollouts per task under the current harness $h_k$
\[
  y_k^{(i,j)} \sim \pi_\theta\big(\cdot \mid x^{(i)};\, h_k\big),
  \qquad i = 1, \dots, n, \quad j = 1, \dots, m.
\]
Each round then contributes one piece of evidence
\[
  e_k = \big(h_k,\ \{y_k^{(i,j)}\}_{i,j}\big),
  \qquad
  \mathcal{C}_k = \mathcal{C}_{k-1} \cup \{e_k\},
\]
recording the harness and the rollouts it produced across the batch. When unit test cases are available for each task, the evidence also carries the outcome of every rollout, so that
\[
  e_k = \big(h_k,\ \{y_k^{(i,j)}\}_{i,j},\
  \{R(y_k^{(i,j)}, g^{(i)})\}_{i,j}\big).
\]

A meta agent $\mathcal{M}$ drives the search. It reads an observation $\Phi(\mathcal{C}_{k-1})$ of the accumulated store and emits the next harness,
\[
  h_k = \mathcal{M}\big(\Phi(\mathcal{C}_{k-1})\big),
  \qquad k = 2, \dots, K,
\]
starting from a seed harness $h_1$. The search optimizes the harness toward expected success on the task distribution, or on a held-out distribution,
\[
  h^\star = \operatorname*{arg\,max}_{h}\
  \mathbb{E}_{x \sim \mathcal{X}}\big[\, R(y, g) \,\big],
  \qquad y \sim \pi_\theta(\cdot \mid x;\, h).
\]
When no unit test cases are available, we take the most evolved harness, $\hat{h} = h_K$; when unit test cases are available, we select the harness in the store with the highest aggregate outcome, estimated on the task batch or a held-out set,
\[
  \hat{h} = \operatorname*{arg\,max}_{k \le K} \bar{R}(h_k),
  \qquad
  \bar{R}(h_k) = \frac{1}{nm} \sum_{i=1}^{n} \sum_{j=1}^{m}
  R\big(y_k^{(i,j)}, g^{(i)}\big).
\]
We then sample the final trajectory for the target task $x$ under this harness, $\hat{y} \sim \pi_\theta(\cdot \mid x;\, \hat{h})$.

We instantiate harness evolution with AHE~\citep{lin2026agentic}, with its \textit{explore agent} disabled so that improvements come from evolving the harness on feedback rather than from retrieving benchmark-specific harnesses externally. See Appendix~\ref{sec:appendix_tb_experiment_details} for details.

\subsection{Harness Scaling}
\label{sec:method_harness_scaling}
To further investigate whether harness improvements arise from reusable design or from additional test-time computation, we introduce \emph{harness scaling}  (Figure~\ref{fig:framework} (d)), a harness-level analogue of test-time scaling. Whereas harness evolution aims to improve a harness across a set of tasks and produce a reusable design, harness scaling adapts the harness for a single evaluation instance by allocating additional inference budget to harness revision. In this sense, harness scaling can be viewed as instance-guided harness adaptation, in contrast to dataset-guided harness evolution. From an initial harness $h_1$, we sample for each task
\[
  y_1 \sim \pi_\theta(\cdot \mid x; h_1).
\]
After observing each trajectory, the agent updates the harness through a meta agent $\mathcal{M}$ instantiated by the same underlying model, and then resamples under the updated harness. When no unit test case is available, the update conditions on the task, the previous harness, and the previous trajectory,
\[
  h_k = \mathcal{M}\big(x,\ \Phi(h_{k-1}, y_{k-1})\big),
  \qquad
  y_k \sim \pi_\theta(\cdot \mid x; h_k),
  \qquad k = 2, \dots, K,
\]
and the latest trajectory is returned $\hat{y} = y_K$.

When a unit test case is available, the update also conditions on the outcome of the previous trajectory,
\[
  h_k = \mathcal{M}\big(x,\ \Phi(h_{k-1}, y_{k-1}, R(y_{k-1}, g))\big),
  \qquad
  y_k \sim \pi_\theta(\cdot \mid x; h_k),
  \qquad k = 2, \dots, K,
\]
so that the meta agent can revise the harness against the signal. We return any trajectory accepted by the outcome,
\[
\hat{y} \in \{\, y_k : R(y_k, g) = 1 \,\}.
\]

\section{Experiments}
In this section, we study whether automatic harness evolution outperforms simple test-time discovery under comparable feedback and inference budgets. We examine three settings: without unit test cases (Section~\ref{sec:experiment_without_test_feedback}), with them (Section~\ref{sec:experiment_with_test_feedback}), and a generalization setting with disjoint search and evaluation tasks (Section~\ref{sec:experiment_disjoint_search_eval}). Across all three, we find that harness evolution does not consistently outperform simple test-time discovery, and that its advantage does not generalize.

\subsection{Experimental Setup}

\paragraph{Benchmark and Models} Following prior work~\citep{leeMetaHarnessEndtoEndOptimization2026, lin2026agentic, zhang2026harnessing}, we evaluate on Terminal-Bench~2.1~\citep{merrill2026terminal}, a verified revision of Terminal-Bench~2.0. It repairs 28 of the 89 tasks, correcting external dependency drift, resource budget mismatches, and instructions misaligned with their tests, while keeping the suite at 89 terminal tasks, providing a more accurate assessment of code agents. We experiment with three frontier models: Claude Opus 4.6, GPT-5.4, and GPT-5.4 mini. Unless otherwise noted, every model is run with a maximum generation budget of 128k tokens and a high reasoning effort setting. To reduce variance from stochastic rollouts, all results are averaged over two independent runs.

\vspace{-2mm}
\paragraph{Implementation}
To ensure a fair comparison, both Harness Scaling and Harness Evolution are initialized with the same initial harness used in AHE~\citep{lin2026agentic}, denoted by $h_1$. We also use this harness as the fixed harness $h$ for Parallel Sampling and Sequential Refinement. For AHE, we sample one rollout per task for each harness, setting $m = 1$, to match the rollout budget across methods. We use a compute budget of $K = 5$ across methods. 

\subsection{Without Unit Test Cases, Automatic Harness Evolution Underperforms Test-Time Scaling}
\label{sec:experiment_without_test_feedback}

\begin{table*}[h]
\centering
\resizebox{0.95\linewidth}{!}{
\begin{tabular}{lccc|c}
\toprule[1pt]
 \textbf{Method} & \textbf{Claude Opus 4.6} & \textbf{GPT-5.4} & \textbf{GPT-5.4 mini} & Average\\
\midrule
direct sampling using initial harness &69.9 &75.3 &59.4 &68.2 \\
\addlinespace[2pt]
\hdashline
\addlinespace[2pt]
\textbf{Test-Time Scaling} & & & & \\
\hspace{2em} Parallel Sampling &\underline{74.7} &\textbf{79.2} &\textbf{62.9} &\textbf{72.3}\\
\hspace{2em} Sequential Refinement &73.0 &73.0 &\underline{61.8} &69.3\\
\addlinespace[2pt]
\hdashline
\addlinespace[2pt]
\textbf{Automatic Harness Evolution} & & & & \\
\hspace{2em} Harness Evolution &71.4 &69.7 &61.3 &67.4\\
\hspace{2em} Harness Scaling &\textbf{76.0} &\underline{78.1} & 61.2 &\underline{71.8}\\
\bottomrule[1pt]
\end{tabular}
}
\caption{Experimental results on Terminal-Bench 2.1 when unit test cases are unavailable. All results report pass@1. Best results in \textbf{bold} and second best in \underline{underline}. Under this setting, automatic harness evolution algorithms underperform simple test-time scaling methods.}
\label{tab:exp1}
\vspace{-2mm}
\end{table*}

We first present results for the setting in which unit test cases are unavailable. In this setting, the agent must rely solely on its own generated trajectories to revise either its solution trajectory or its harness. For parallel methods, the final answer is selected by a self judge over multiple sampled candidates. For sequential methods, the final answer is taken from the last refinement step. This setting therefore tests whether the agent can improve through self evaluation alone, rather than through explicit correctness signals.

Table~\ref{tab:exp1} shows the results. Overall, automatic harness evolution does not deliver a stronger performance gain than test-time scaling methods, and its effectiveness varies substantially across models. 
Parallel Sampling is the most consistent approach, improving the average score from 68.2 to 72.3 and yielding gains on all three models. Sequential Refinement provides only a marginal average improvement of 1.1 points and even slightly degrades GPT-5.4 relative to direct sampling.
In contrast, Harness Evolution fails to outperform the direct sampling baseline on average. The degradation is most pronounced on GPT-5.4, where performance drops sharply from 75.3 to 69.7, indicating that iterative harness revision can actively hurt a strong model when the revision process is guided only by the agent's own judgment. Harness Scaling partially closes this gap, reaching the best single result on Claude Opus 4.6 and a competitive average of 71.8. 

Taken together, these results suggest that without unit test cases, even strong agents struggle to extract reliable learning signals from their own trajectories.
Self-generated feedback is noisy, and sequential revision based on such feedback risks compounding early mistakes. Simpler parallel sampling proves more effective in this feedback limited setting. This indicates that harness revision  likely requires a reliable external correctness signal to ground the revision process. 

\subsection{With Unit Test Cases, Automatic Harness Evolution Underperforms Test-Time Scaling}

\label{sec:experiment_with_test_feedback}

\begin{table*}[h]
\centering
\resizebox{\linewidth}{!}{
\begin{tabular}{lcccccc|cc}
\toprule[1pt]
& \multicolumn{2}{c}{\textbf{Claude Opus 4.6}}  & \multicolumn{2}{c}{\textbf{GPT-5.4}} & \multicolumn{2}{c}{\textbf{GPT-5.4 mini}} & \multicolumn{2}{c}{Average}\\
\cmidrule(lr){2-3} \cmidrule(lr){4-5} \cmidrule(lr){6-7} \cmidrule(lr){8-9}
\textbf{Method} & \textbf{pass@1} & \textbf{pass@5} & \textbf{pass@1} & \textbf{pass@5} & \textbf{pass@1} & \textbf{pass@5} & \textbf{pass@1} & \textbf{pass@5} \\
\midrule
direct sampling using initial harness &69.9 &- &75.9 &- &56.9 &- &67.6 &- \\
\addlinespace[2pt]
\hdashline
\addlinespace[2pt]
\textbf{Test-Time Scaling} & & & & & & & &\\
\hspace{2em} Parallel Sampling &\textbf{84.8} &84.8 &\textbf{87.1} &87.1 &\textbf{74.1} &74.1 &\textbf{82.0} &82.0 \\
\hspace{2em} Sequential Refinement &\underline{83.1} &\textbf{90.4} &\underline{85.4} &\textbf{93.3} &\underline{62.1} &\underline{77.6} &\underline{76.9} &\textbf{87.1} \\
\addlinespace[2pt]
\hdashline
\addlinespace[2pt]
\textbf{Automatic Harness Evolution} & & & & & & & & \\
\hspace{2em} Harness Evolution &73.0 &83.2 &78.6 &\underline{89.3} &57.4 &64.6 &69.7 &79.0 \\
\hspace{2em} Harness Scaling &\underline{83.1} &\underline{89.9} &82.0 &88.8 &61.5 &\textbf{78.2} &75.5 &\underline{85.6} \\
\bottomrule[1pt]
\end{tabular}
}
\caption{Experimental results on Terminal-Bench 2.1 when unit test cases are available. Best results in \textbf{bold} and second best in \underline{underline}. Automatic harness evolution algorithms still underperform test-time scaling baselines in this setting, on both pass@1 and pass@5.}
\label{tab:exp2}
\end{table*}

Second, we compare all methods when unit test cases are available. The unit test cases are used both as feedback for iterative refinement and as an oracle for selecting the final trajectory. 
As unit tests allow the best of several sampled trajectories to be identified, we report both pass@1 and pass@5.


As shown in Table~\ref{tab:exp2}, we observe a similar trend under this setting. All test-time scaling and automatic harness evolution methods substantially improve over direct sampling with the initial harness, confirming that unit test feedback helps regardless of the mechanism. However, neither Harness Evolution nor Harness Scaling outperforms the simpler baselines on either metric. On pass@1, Harness Evolution barely improves over direct sampling, while Parallel Sampling achieves the best average of 82.0. On pass@5, Sequential Refinement remains the strongest method with an average of 87.1, surpassing Harness Evolution (79.0) and Harness Scaling (85.6).

If harness revision genuinely produced better harnesses, we would expect the improvement to be reflected in pass@1. Instead, the benefit only materializes when we can select among multiple trajectories. This suggests that existing harness evolution algorithms do not reliably enable the agent to solve previously unsolved tasks through improved harness design. Rather, their gains largely stem from making multiple attempts, an effect that Parallel Sampling and Sequential Refinement achieve more directly and more effectively. This further indicates that refining solutions is a more productive use of extra compute than revising the harness itself for Terminal-Bench tasks.

\subsection{Harness Evolution Fails to Generalize Beyond Training Tasks}
\label{sec:experiment_disjoint_search_eval}

\newcommand{\gain}[1]{\rlap{\,{\scriptsize\textcolor{gray}{(#1)}}}}

\begin{table*}[h!]
\centering
\resizebox{0.95\linewidth}{!}{
\begin{tabular}{lccc@{\hspace{2em}}|c@{\hspace{2em}}}
\toprule[1pt]
\textbf{Method} & \textbf{Claude Opus 4.6}  & \textbf{GPT-5.4} & \textbf{GPT-5.4 mini} & Average\\
\midrule
direct sampling using initial harness &63.3 &72.1 &54.4 &63.2\\
\addlinespace[2pt]
\hdashline
\addlinespace[2pt]
Harness Evolution &64.5\gain{+1.2} &72.1\gain{+0.0} &53.0\gain{-1.4} &63.2\gain{+0.0} \\
\bottomrule[1pt]
\end{tabular}
}
\caption{Experimental results on Terminal-Bench 2.1 with disjoint search and evaluation tasks. We perform Harness Evolution on a training set with unit test cases, select the resulting harness based on validation performance, and report pass@1 on the test set. Compared with the pass@1 test results using the initial harness, Harness Evolution yields no significant gains, suggesting limited generalization and a strong tendency to overfit the search set.}
\label{tab:exp3}
\end{table*}


Third, we evaluate the generalization ability of Harness Evolution, namely whether a harness evolved on one set of tasks can transfer to unseen tasks. Specifically, we split Terminal-Bench 2.1 into 45 training tasks, 10 validation tasks, and 34 held-out test tasks. We run Harness Evolution on the training set, select the best evolved harness based on validation performance, and evaluate it on the test set. The other three methods are not directly applicable in this setting, since they rely on instance-level scaling or refinement rather than producing a reusable artifact.


As shown in Table~\ref{tab:exp3}, the evolved harness yields no meaningful improvement over the initial harness on the test set: it gains a mere 1.2 points on Claude Opus 4.6, leaves GPT-5.4 unchanged, and in fact costs 1.4 points on GPT-5.4 mini, so the average across models is exactly flat. This stands in sharp contrast to the improvements Harness Evolution achieves when evaluated on the same tasks it is optimized on, as reported in the previous experiments. The discrepancy suggests that the revisions discovered during evolution encode task-specific shortcuts rather than genuinely better harness design principles. In other words, based on the Terminal-Bench results, current harness evolution algorithms exhibit limited generalization ability and appear prone to severe overfitting to the training tasks, especially for weaker models. 

\section{Harness Evolution for Long-Horizon Games}\label{sec:long-horizon}

Following prior work, we evaluate all algorithms on Terminal-Bench and find that automatic harness evolution is often less effective than simple methods that scale computation at test time. While this may simply reflect that current models are not yet capable enough to revise the entire harness, in which case stronger priors could help, or they may already be optimized for Terminal-Bench during training, we offer two additional explanations for this result. First, agents already achieve relatively high scores on Terminal-Bench, and the small set of remaining failures may stem from limitations in the underlying model rather than deficiencies in the harness. Second, Terminal-Bench may simply not be very sensitive to harness design: a minimal setup consisting of a shell tool and a basic prompt already suffices for most solvable tasks, so performance is bottlenecked by the model's reasoning rather than by the surrounding scaffolding. Modifying prompts, tool configurations, or middleware therefore yields only marginal gains.


These observations suggest that the value of automatic harness evolution might be demonstrated in settings that are difficult enough to leave current agents substantial headroom for improvement, and that depend heavily on the harness, with specialized tools, skills, or workflows being crucial to success. Motivated by this, we further evaluate on two long-horizon game benchmarks that (1) are sufficiently challenging, (2) lie out of distribution, such that success relies heavily on adaptation, and (3) are designed to provide granular, verifiable feedback from the environment. Experimental results show that, in such settings, automatic harness evolution yields clear benefits.

\subsection{Experimental Setup}

\paragraph{Benchmark and Models} We evaluate on ARC-AGI-3~\citep{arcprize2026arcagi3}, an interactive visual benchmark with turn-based environments in which agents must explore, infer goals, and plan effective action sequences without explicit instructions, and \textit{Interactive Games \& Simulators} tasks of EdgeBench~\citep{zhu2026edgebench}, which feature long-horizon agent operation with rich, multilevel feedback. Both benchmarks pose significant challenges even for frontier models and demand substantial adaptation to novel environments. We experiment with Claude Opus 4.8 and GPT models (GPT-5.6 Terra on ARC-AGI-3 and GPT-5.4 on EdgeBench).

\paragraph{Implementation}
Rather than initializing from a minimal harness, we initialize harness evolution from well-engineered harnesses this time. This lets us test whether automatic harness evolution can improve on a state-of-the-art harness, rather than merely recovering design choices that are already standard practice. Specific choices are detailed in the following sections. 


By task design, the goal is to make use of the rich environmental feedback and optimize performance on each individual task rather than to find a single harness that transfers. We therefore compare two task-level algorithms, Sequential Refinement and Harness Scaling, under the same budget.
Throughout the run, the agent receives continuous, fine-grained feedback (e.g., pass rates, failing tests, and scores) and iteratively refines its solution or harness in a closed loop until the budget is exhausted. This feedback also guards against regression, so we compare the best score achieved throughout the entire trajectory.
More details can be found in Appendices~\ref{sec:appendix_arcagi_experimental_details} and \ref{sec:appendix_edgebench_experimental_details}.


\subsection{ARC-AGI-3}
\label{sec:arcagi}
\begin{table*}[h!]
\centering
\setlength{\tabcolsep}{4pt}
\resizebox{1.0\linewidth}{!}{
\begin{tabular}{lcccc|cccc}
\toprule[1pt]
 & \multicolumn{4}{c|}{\textbf{Claude Opus 4.8}} & \multicolumn{4}{c}{\textbf{GPT-5.6 Terra}}\\
 \cmidrule[0.75pt](lr){2-5} \cmidrule[0.75pt](lr){6-9}
 & \multicolumn{2}{c}{\textbf{Sequential Refinement}} & \multicolumn{2}{c|}{\textbf{Harness Scaling}} & \multicolumn{2}{c}{\textbf{Sequential Refinement}} & \multicolumn{2}{c}{\textbf{Harness Scaling}} \\
  \textbf{Task} & [Claude Code] & [VISTA*] & [VISTA] & [full harness] & [Codex] & [VISTA*] & [VISTA] & [full harness]\\
\midrule
ar25  &41.7	&\underline{58.3}	&\textbf{100}	&\textbf{100} &5.8	&29.1	&\underline{89.8}	&\textbf{100}\\
\addlinespace[1pt]
bp35  &0.8	&\underline{5.4}	&5.3	&\textbf{8.8} &0.6	&\underline{2.4}	&1.7	&\textbf{6.6} \\
\addlinespace[1pt]
cd82  &10.5	&\textbf{100}	&\underline{90.6}	&\textbf{100} &0.0	&39.3	&\underline{65.5}	&\textbf{93.4} \\
\addlinespace[1pt]
cn04  &\underline{47.6}	&\underline{47.6}	&\underline{47.6}	&\textbf{71.4} &14.3	&38.8	&\underline{45.4}	&\textbf{47.6} \\
\addlinespace[1pt]
dc22  &14.3	&23.6	&\textbf{47.6}	&\underline{44.4} &4.8	&\textbf{34.2}	&\underline{30.6}	&28.2 \\
\addlinespace[1pt]
ft09  &\textbf{100}	&\textbf{100}	&\textbf{100}	&\textbf{100} &22.3	&83.2	&\textbf{90.6}	&\underline{87.1} \\
\addlinespace[1pt]
g50t  &3.6 &1.8 &\underline{41.6} &\textbf{48.7} &\textbf{10.7} &\underline{2.7} &1.8 &\underline{2.7} \\
\addlinespace[1pt]
ka59  &53.6&44.7&\underline{89.9}&\textbf{100} &0.3 &20.6 &\underline{21.5} &\textbf{45.8} \\
\addlinespace[1pt]
lf52  &4.6&10.9&\underline{18.2}&\textbf{32.5} &3.9 &\underline{8.4} &8.0 &\textbf{11.5} \\
\addlinespace[1pt]
lp85  &16.7 &\underline{72.5} &\textbf{100} &\textbf{100} &2.8 &5.7 &\underline{9.0} &\textbf{38.9} \\
\addlinespace[1pt]
ls20  &33.4&12.7&\underline{83.4}&\textbf{100} &2.8 &\underline{4.7} &4.5 &\textbf{8.3} \\
\addlinespace[1pt]
m0r0  &\textbf{100}&\underline{93.4}&\textbf{100}&\textbf{100} &14.3 &\underline{71.8} &8.8 &\textbf{100} \\
\addlinespace[1pt]
r11l  &47.6&\textbf{100}&\textbf{100}&\underline{89.4} &4.8 &\underline{8.8} &\textbf{31.8} &5.5 \\
\addlinespace[1pt]
re86  &58.3&41.7&\underline{77.8}&\textbf{100} &8.3 &16.2 &\underline{25.0} &\textbf{44.6} \\
\addlinespace[1pt]
s5i5  &\underline{41.3} &3.0 &20.8 &\textbf{100} &1.6 &0.6 &\underline{15.2} &\textbf{30.0} \\
\addlinespace[1pt]
sb26  &95.3&\textbf{100}&\textbf{100}&\underline{97.4} &0.0 &\textbf{100} &66.6 &\underline{74.3} \\
\addlinespace[1pt]
sc25  &\textbf{100}&42.1&\underline{58.6}&\textbf{100} &0.0 &\underline{24.7} &17.4 &\textbf{40.0} \\
\addlinespace[1pt]
sk48  &\textbf{24.0}&8.3&\underline{23.3}&14.0 &0.0 &\textbf{0.8} &\underline{0.3} &0.0 \\
\addlinespace[1pt]
sp80  &0.3&\underline{47.6}&4.8&\textbf{99.5} &0.0 &\underline{1.6} &1.1 &\textbf{4.8} \\
\addlinespace[1pt]
su15  &13.3&\underline{22.2}&6.7&\textbf{79.7} &11.8 &9.1 &\textbf{48.4} &\underline{48.2} \\
\addlinespace[1pt]
tn36  &0.0&\textbf{18.5}&3.0&\underline{3.6} &0.0 &\underline{9.4} &\textbf{27.9} &4.1\\
\addlinespace[1pt]
tr87  &\underline{71.4}&0.6&\underline{71.4}&\textbf{100} &0.0 &\underline{11.8} &1.5 &\textbf{100}\\
\addlinespace[1pt]
tu93  &\underline{93.2}&\textbf{100}&\textbf{100}&91.9 &6.7&3.2&\textbf{26.4}&\underline{25.5}\\
\addlinespace[1pt]
vc33  &\underline{75.0}&14.2&30.0&\textbf{100} &0.0&\underline{0.6}&\textbf{1.2}&0.0\\
\addlinespace[1pt]
wa30  &\textbf{62.2}&13.3&\textbf{62.2}&\underline{55.7} &0.0&0.1&\underline{6.7}&\textbf{6.8}\\
\addlinespace[2pt]
\hdashline
\addlinespace[3pt]
\textbf{Average} &44.3 &43.3 &\underline{59.3} &\textbf{77.5} &4.6 &21.1 &\underline{25.9} &\textbf{38.2} \\
\bottomrule[1pt]
\end{tabular}
}
\caption{Results on all 25 public ARC-AGI-3 games, reported as official RHAE scores (win rates in Table~\ref{tab:arcagi_winrate}). Best results in \textbf{bold}, second best \underline{underlined}. [VISTA*] disables memory evolution; [VISTA] is the original implementation with memory evolution; [full harness] additionally evolves skills, tools, and hooks. Automatic harness evolution yields clear benefits in this setting.}
\label{tab:arcagi}
\end{table*}

We first evaluate on ARC-AGI-3, a suite of interactive visual games in which an agent must discover the rules and goals of unfamiliar worlds through interaction. The benchmark scores agents using Relative Human Action Efficiency (RHAE), which measures per-level action efficiency against a human baseline, normalized per game, so that both completion and efficiency are captured.

We initialize evolution from VISTA*, a plain version of VISTA~\citep{vista2026} with its memory-evolution components disabled. VISTA is a strong visual harness that reports perfect RHAE scores on all 25 public games using Claude Opus 5.0. We evaluate two variants of Harness Scaling: one that evolves only memory, following the original VISTA implementation, and one that is allowed to evolve the full harness. In both cases we follow VISTA in adopting a \textit{free} update mechanism, in which harness updates are exposed as tools (e.g., \texttt{read\_skill} and \texttt{write\_skill}) that the agent may call at any point during the run. We set the budget to 2,000 actions and also report the performance of plain Claude Code CLI and Codex CLI for reference.



We experiment with Claude Opus 4.8 and GPT-5.6 Terra, which are strong enough to make progress on these games but not so strong that the tasks are saturated, allowing differences between methods to remain visible. Table~\ref{tab:arcagi} reports the full results. Harness Scaling substantially outperforms Sequential Refinement under both models. The [full harness] variant improves over [VISTA*] from 43.3 to 77.5 RHAE with Claude Opus 4.8 (+79.0\% relative) and from 21.1 to 38.2 with GPT-5.6 Terra (+81.0\%). These gains reflect both an expanded set of solvable problems and improved efficiency.

Comparing [VISTA] with [full harness] isolates the contribution of skills and tools beyond memory, and it is substantial: 59.3 to 77.5 with Claude Opus 4.8 and 25.9 to 38.2 with GPT-5.6 Terra. For GPT-5.6 Terra the effect is especially pronounced, as memory evolution alone adds only 4.8 points over [VISTA*] while skills and tools add a further 12.3. The agents write tools that guide exploration and game understanding, such as \texttt{frame\_diff}, \texttt{link\_states}, and \texttt{route\_markers}, and in several cases these unlock games that no other configuration solves: on \texttt{tr87}, for instance, GPT-5.6 Terra goes from 11.8 to 100 RHAE. See Appendices~\ref{sec:appendix_arcagi_artifacts} and~\ref{sec:appendix_arcagi_casestudy} for examples.

\subsection{EdgeBench} \label{sec:edgebench}
Next, we apply Harness Scaling to the \textit{Interactive Games \& Simulators} tasks of EdgeBench, where agents face strong out-of-distribution pressure and must develop and refine strategies through high-frequency interaction across many episodes. We use Claude Code CLI as the initial harness for the Claude model and Codex CLI as the initial harness for the GPT model. These vendor-developed harnesses are extensively engineered and co-designed with their respective models, so they serve as strong baselines. We fix the budget to a 12-hour wall-clock window and implement a \textit{periodic} harness update mechanism under Harness Scaling: a harness update is triggered whenever the agent has made three submissions or 120 minutes have elapsed since the last update, whichever comes first. This ensures the harness is revisited at regular intervals throughout the run. At each update, the meta agent may also choose to leave the harness unchanged.

\begin{table*}[h!]
\centering
\resizebox{1.0\linewidth}{!}{
\begin{tabular}{lcccc}
\toprule[1pt]
 & \multicolumn{2}{c}{\textbf{Claude Opus 4.8}} & \multicolumn{2}{c}{\textbf{GPT-5.4}}\\
 \cmidrule[0.75pt](lr){2-3} \cmidrule[0.75pt](lr){4-5}
 \textbf{Task} & \textbf{Sequential Refinement} & \textbf{Harness Scaling} & \textbf{Sequential Refinement} & \textbf{Harness Scaling}\\
\midrule
dcss\_dungeon\_ai  &17.1 &\textbf{22.4}  &9.9 &\textbf{13.3} \\
\addlinespace[1pt]
nethack\_dungeon\_agent  &44.3 &\textbf{61.3} &\textbf{50.4} &49.0 \\
\addlinespace[1pt]
openrct2\_theme\_park\_ai &38.4 &\textbf{47.8} &36.2 &\textbf{50.0}\\
\addlinespace[1pt]
openttd\_transport\_ai &53.5 &\textbf{56.2} &\textbf{51.6} &50.5\\
\addlinespace[1pt]
wesnoth\_tactical\_ai &\textbf{88.0} &86.0 &81.5 &\textbf{86.0}\\
\addlinespace[2pt]
\hdashline
\addlinespace[2pt]
\textbf{Average} & 48.3 & \textbf{54.7} & 45.9 & \textbf{49.8} \\
\bottomrule[1pt]
\end{tabular}
}
\caption{Experimental results on \textit{Interactive Games \& Simulators} tasks of EdgeBench. The highest score achieved during the 12-hour exploration period is reported. All results are averaged over two independent runs. Automatic harness evolution (Harness Scaling) shows better performance than test-time scaling (Sequential Refinement) under this setting.}
\label{tab:edgebench}
\end{table*}


As shown in Table~\ref{tab:edgebench}, automatic harness evolution outperforms test-time scaling by 13.3\% on average with Claude Opus 4.8 and by 8.5\% with GPT-5.4, demonstrating that it can be beneficial on long-horizon tasks with substantial headroom and rich feedback. Qualitatively, memory remains the dominant interface through which the harness is modified. The meta agent helps by diagnosing failures, breaking redundant loops, and flagging overlooked details, and it is most effective on tasks requiring deep exploration along a single path; it sometimes neglects breadth, failing to prompt exploration of alternative directions. The two models also differ in style: Claude Opus 4.8 writes concise verdicts and steers the agent during execution, while GPT-5.4 favors writing plans.

\begin{section}{Conclusions}\label{sec:conclusion}
We study how automatic harness evolution for LLM agents is evaluated and observe that current methods commonly omit comparison with test-time scaling algorithms and use feedback from unit test cases to search, then report final performance on the same benchmark, conflating genuine advances in harness design with adaptation to the specific evaluation tasks.
Experiments on Terminal-Bench show that it fails to outperform test-time scaling baselines under a matched budget, and that its gains generalize poorly beyond the search tasks.
Further experiments show that long-horizon games are a promising setting for automatic harness evolution, as they are difficult enough, rely heavily on adaptation to out-of-distribution dynamics, and provide granular feedback by design.
These findings suggest that current evidence for automatic harness evolution should be interpreted with caution, and highlight the need to identify the settings in which it is genuinely helpful.

\end{section}

\bibliography{iclr2027_conference}
\bibliographystyle{iclr2027_conference}

\newpage
\appendix
\section{Experiments on Terminal-Bench~2.1}
\label{sec:appendix_tb}

\subsection{Experimental Details}
\label{sec:appendix_tb_experiment_details}


\paragraph{Initial Harness}
Following prior work, all four methods initialize the code agent from an identical minimal harness: the model is granted access to a single bash tool and nothing else, without skills, middleware, or persistent memory.

\paragraph{Summarization Map}
To distill actionable evidence from raw rollout trajectories, we employ the Agent Debugger, a summarization agent introduced in the AHE framework~\citep{lin2026agentic}. Given multiple traces collected for each task, the Agent Debugger frames the trajectories as a navigable environment of files, where each trajectory message resides in its own file and is explored through generic shell and scripting tools. Traces that share the same query are grouped into a single environment, and the debugger is tasked with identifying the root cause of each failure or the pattern behind each success. Its findings are written to an analysis report for every task, annotated with pass or fail status, and then aggregated into a benchmark level overview that serves as an entry point for downstream analysis. The original traces remain available in both raw and lightly processed form, so claims in the reports can be verified on demand. All of this content is exposed as files that support progressive disclosure, which reduces token consumption while preserving access to full detail when needed.

\paragraph{Agentic Harness Engineering (AHE)}

AHE ordinarily couples two components: an explore agent that retrieves harnesses tuned for the benchmark under evaluation from external sources, and an evolution loop that proposes and refines the harness from the feedback collected on the benchmark itself. Because the explore agent imports benchmark-specific harnesses rather than discovering improvements from feedback, it conflates reusable harness evolution with retrieval of solutions already fitted to the evaluation tasks. We therefore disable it and retain only the propose and refine loop, so that every reported gain is attributable to evolution over the feedback signal rather than to externally sourced, benchmark-specific harnesses.

\paragraph{Configuration}
The full configuration used throughout is listed in
Table~\ref{tab:hparams}; these values come from a snapshot of the reference run and remain fixed in every experiment. Each rollout executes in its own freshly provisioned E2B remote sandbox.

\begin{table}[h!]
    \centering
    \small
    \setlength{\tabcolsep}{6pt}
    \renewcommand{\arraystretch}{1.05}
    \begin{tabular}{lll}
        \toprule[1pt]
        \textbf{Agent} & \textbf{Hyperparameter} & \textbf{Value} \\
        \midrule
        Code Agent
            & Model              & Claude Opus 4.6 / GPT-5.4 / GPT-5.4 mini \\
            & Reasoning effort           & high \\
            & Max context                & 200{,}000 tokens \\
            & Max generation per turn    & 128{,}000 tokens \\
            & Max model turns            & 300 \\
        \midrule
        Agent Debugger
            & Model              & Claude Opus 4.6 / GPT-5.4 / GPT-5.4 mini \\
            & Reasoning effort           & high \\
            & Max context                & 200{,}000 tokens \\
            & Max generation per turn    & 128{,}000 tokens \\
            & Max model turns            & 25 \\
            & Context compaction threshold & 0.75 \\ 
            & Retry attempts             & 3 \\
            & Retry backoff              & 2.0\,s \\
        \midrule
        Meta Agent
            & Model              & Claude Opus 4.6 / GPT-5.4 / GPT-5.4 mini \\
            & Reasoning effort           & high \\
            & Max context                & 200{,}000 tokens \\
            & Max generation per turn    & 128{,}000 tokens \\
            & Max model turns            & 500 \\
            & Context compaction threshold & 0.75 \\ 
        \bottomrule[1pt]
    \end{tabular}
    \caption{Experiment configurations.}
    \label{tab:hparams}
\end{table}
 
\paragraph{Evaluation Metrics}
\begin{itemize}
    \item \textbf{pass@1} Our primary metric is pass@1. Let $\mathcal{D}$ denote the task set with $k$ rollouts per task, and let $r_{i,j} \in \{0, 1\}$ denote the binary reward of rollout $j$ on task $i$. The pass@1 score is the average reward over all rollouts:
    \begin{equation}
        \mathrm{pass@1}
        \;=\;
        \frac{1}{k\,\lvert \mathcal{D} \rvert}
        \sum_{i=1}^{\lvert \mathcal{D} \rvert}
        \sum_{j=1}^{k}
        r_{i,j}.
        \label{eq:pass1}
    \end{equation}
    Rollouts that terminate on an infrastructure exception, such as a sandbox crash or an API timeout, are scored as $r = 0$ rather than excluded from the average.

    \item \textbf{pass@k}
    To measure how much repeated sampling helps, we additionally report pass@$k$, the fraction of tasks solved by at least one of the $k$ rollouts, in the setting where unit test cases are available (Section~\ref{sec:experiment_with_test_feedback}):
    \begin{equation}
        \mathrm{pass@}k
        \;=\;
        \frac{1}{\lvert \mathcal{D} \rvert}
        \sum_{i=1}^{\lvert \mathcal{D} \rvert}
        \max_{j \in \{1, \dots, k\}} r_{i,j}.
        \label{eq:passk}
    \end{equation}
    The same scoring convention applies: rollouts lost to infrastructure exceptions count as failures. Since a single success suffices for a task to count as solved, pass@5 upper bounds pass@1, and the gap between the two reflects the variance of the agent across independent attempts.
\end{itemize}

\paragraph{Cost}
In our cost estimation, we did not include the cost of modifying the harness. This is because it is very cheap compared with generating a new trajectory. Moreover, given our conclusion that automatic harness evolution (which incurs additional cost for harness modification) underperforms test-time scaling under comparable budgets, counting this cost would only strengthen our conclusion.

\subsection{Rational Harness Edits but Marginal Gains}
To understand where the gains and limits of automatic harness evolution come from, we take a closer look at the trajectories produced by Harness Evolution and Harness Scaling on Terminal-Bench. Specifically, we examine what the meta agent modifies at each iteration, which changes are kept or rolled back, and which failure classes the changes target. We find that while the meta agent makes rational, well-motivated edits across the prompt, middleware, and tool layers, a stable core of hard tasks remains unaffected, and the resulting improvements stay limited beyond simple test-time discovery baselines.

\paragraph{Harness Evolution}
Under Harness Evolution, the meta agent's first move is usually at the prompt layer, adding behavioral rules to the system prompt and long-term memory that target recurring failure classes, such as instructions to produce deliverables early with greater budget awareness, to copy fragile state before mutating it, and to recheck task constraints before finishing. When such advisory text plateaus, the meta agent escalates to runtime enforcement through middleware, such as turn budget trackers that remind the agent to ship output once turn or time thresholds are reached, truncation of oversized tool outputs, and finalization gates that block completion when deliverables are missing or unverified. It also edits the tool layer, correcting misleading tool guidance and injecting recovery hints at command execution time. 

\paragraph{Harness Scaling} 
The dominant modification pattern in Harness Scaling is to encode knowledge about specific failures from previous rollouts into the next iteration's prompt or long-term memory. For example, the meta agent records known bugs, implementation templates, verification checks, file paths, and command sequences for particular tasks. It also frequently improves workflow efficiency. Many changes replace repeated polling, exploratory commands, or slow setup sequences with compact command plans: batching installation and verification commands, skipping unnecessary environment detection, allowing longer shell timeouts for slow package installations, and prescribing workflows with explicit message budgets. These changes are especially helpful for tasks where the agent already knew the correct high level solution but lost time or failed due to timeouts, excessive polling, or incomplete setup. Another common class of edits strengthens verification behavior. The meta agent adds syntax checks, unit tests, or local checks that mimic the verifier before final submission. 

Despite the breadth of these modifications, the overall improvement remains limited relative to test-time discovery baselines. The core issue is that most edits memorize fixes rather than distilling strategies. Much of this information is precisely what a competent agent can rediscover through exploration within a single rollout, so persisting it in the harness saves time on tasks the agent could already solve but rarely converts failures into successes. Meanwhile, the stable core of hard failures, which stem from deep domain reasoning demands or constraints outside harness control, remains unaffected by accumulated knowledge, and the growing volume of persistent prompt text introduces context bloat that can offset the remaining gains.

\begin{figure*}[h!]
\centering
\includegraphics[width=1.0\linewidth]{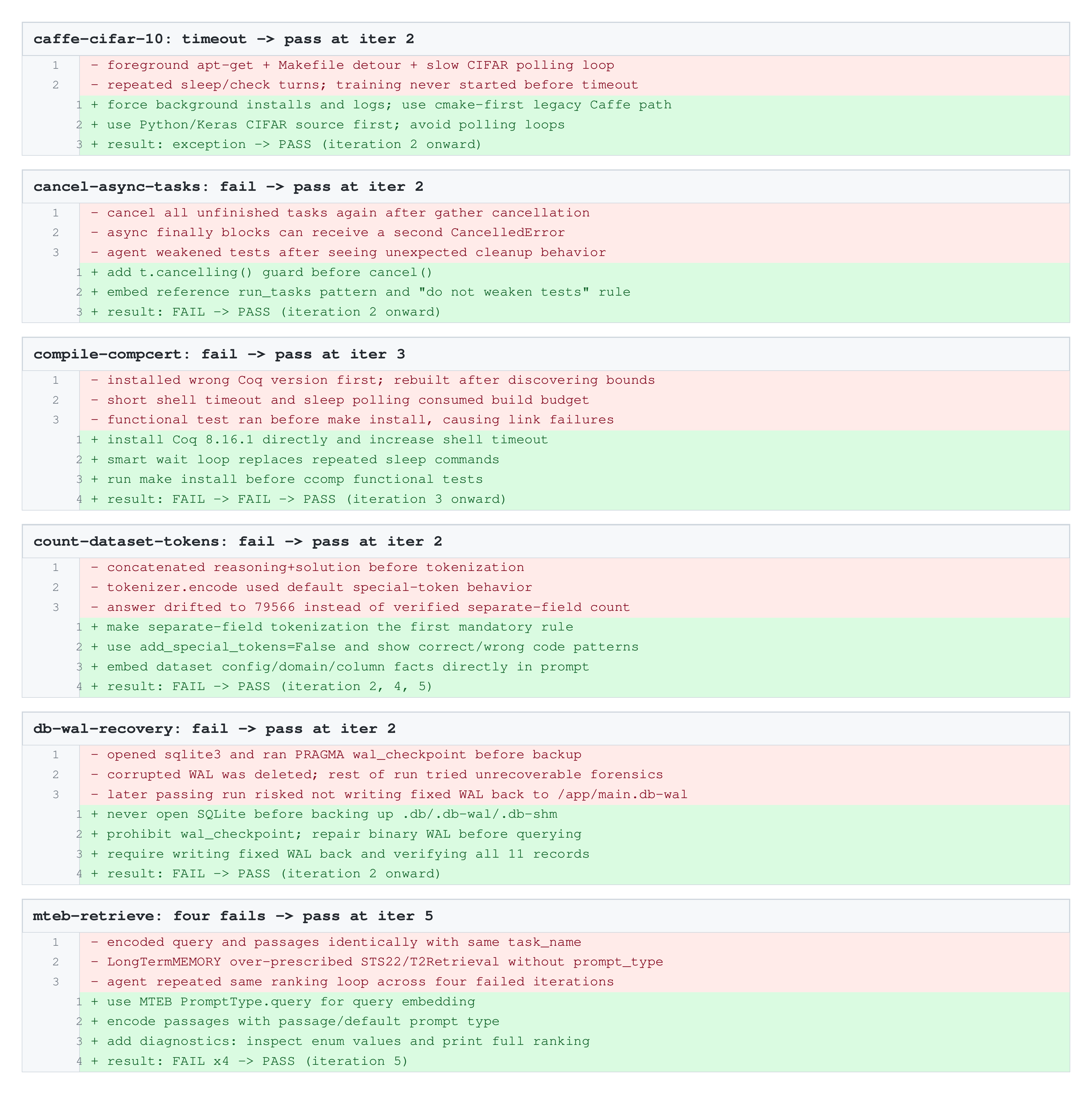}
\caption{Examples of harness modifications made in Harness Scaling.}
\label{fig:case_harness_scaling}
\end{figure*}

Figure~\ref{fig:case_harness_scaling} presents representative examples of the modifications the meta agent applies to the harness. These modifications target recurring weaknesses observed in failed rollouts, including fragile setup and build procedures, wasteful polling loops and short shell timeouts, unsafe cleanup behavior, unpinned dependency versions, incorrect assumptions about data formats, destructive inspection of state that must be preserved, misconfigured embedding prompt types, and attempts by the agent to weaken tests. Notably, many fixes take the form of task-specific rules and facts embedded directly into the prompt, such as prescribed command orderings and verified dataset properties, echoing our earlier observation that edits tend to memorize fixes rather than distill general strategies. However, although these edits are individually reasonable, the stable core of hard failures remains unaffected.

\subsection{Compute Budget $K$}
\label{sec:appendix_tb_ablation}
We additionally conduct an ablation on the compute budget $K$ to test whether our conclusions hold under a larger budget. Specifically, we double the budget to $K = 10$ and rerun the experiment under the same setting with Claude Opus 4.6, the model on which automatic harness evolution performs best. All results are averaged over two independent runs.

\begin{table*}[h]
\centering
\resizebox{0.85\linewidth}{!}{
\begin{tabular}{lcc}
\toprule[1pt]
\textbf{Method} & \textbf{Claude Opus 4.6} [$K=5$] & \textbf{Claude Opus 4.6} [$K=10$] \\
\midrule
direct sampling using initial harness &69.9 &69.5 \\
\addlinespace[2pt]
\hdashline
\addlinespace[2pt]
\textbf{Test-Time Scaling} & & \\
\hspace{2em} Parallel Sampling &\underline{74.7} &\textbf{73.6} \\
\hspace{2em} Sequential Refinement &73.0 &71.6 \\
\addlinespace[2pt]
\hdashline
\addlinespace[2pt]
\textbf{Automatic Harness Evolution} & & \\
\hspace{2em} Harness Evolution &71.4 &\underline{71.9} \\
\hspace{2em} Harness Scaling &\textbf{76.0} &\textbf{73.6} \\
\bottomrule[1pt]
\end{tabular}
}
\caption{Experimental results on Terminal-Bench 2.1 when unit test cases are unavailable. All results report pass@1. Best results in \textbf{bold} and second best in \underline{underline}. Even with a larger compute budget, we find no clear evidence that automatic harness evolution outperforms test-time scaling methods. A larger budget does not translate into better performance, as the self-generated feedback is too noisy to guide the search reliably.}
\label{tab:exp1_k10}
\end{table*}

\begin{table*}[h]
\centering
\resizebox{0.9\linewidth}{!}{
\begin{tabular}{lcccc}
\toprule[1pt]
 & \multicolumn{2}{c}{\textbf{Claude Opus 4.6} [$K=5$]}
 & \multicolumn{2}{c}{\textbf{Claude Opus 4.6} [$K=10$]}\\
 \cmidrule(lr){2-3} \cmidrule(lr){4-5}
\textbf{Method} 
&\textbf{pass@1} &\textbf{pass@5} &\textbf{pass@1} &\textbf{pass@10} \\
\midrule
direct sampling using initial harness  &69.9 &- &68.5 &-\\
\addlinespace[2pt]
\hdashline
\addlinespace[2pt]
\textbf{Test-Time Scaling}  & & & &\\
\hspace{2em} Parallel Sampling &\textbf{84.8} &84.8 &88.7 &88.7\\
\hspace{2em} Sequential Refinement  &\underline{83.1} &\textbf{90.4} &\underline{89.6} &\underline{93.1}\\
\addlinespace[2pt]
\hdashline
\addlinespace[2pt]
\textbf{Automatic Harness Evolution}  & & & &\\
\hspace{2em} Harness Evolution  &73.0 &83.2 &73.6 &86.5\\
\hspace{2em} Harness Scaling  &\underline{83.1} &\underline{89.9} &\textbf{90.5} &\textbf{94.4}\\
\bottomrule[1pt]
\end{tabular}
}
\caption{Experimental results on Terminal-Bench 2.1 when unit test cases are available. Best results in \textbf{bold} and second best in \underline{underline}. Even on the strongest model, the gap between automatic harness evolution and test-time scaling remains within run-to-run variance.
}
\label{tab:exp2_k10}
\end{table*}

\begin{table*}[h!]
\centering
\resizebox{0.85\linewidth}{!}{
\begin{tabular}{lcc}
\toprule[1pt]
\textbf{Method} & \textbf{Claude Opus 4.6} [$K=5$] & \textbf{Claude Opus 4.6} [$K=10$]\\
\midrule
direct sampling using initial harness &63.3 &63.3\\
\addlinespace[2pt]
\hdashline
\addlinespace[2pt]
Harness Evolution &64.5\gain{+1.2} &63.0\gain{-0.3} \\
\bottomrule[1pt]
\end{tabular}
}
\caption{Experimental results on Terminal-Bench 2.1 with disjoint search and evaluation tasks. We perform Harness Evolution on a training set with unit test cases, select the resulting harness based on validation performance, and report pass@1 on the test set. Compared with the pass@1 test results using the initial harness, Harness Evolution yields no significant gains, suggesting limited generalization and a strong tendency to overfit the search set.}
\label{tab:exp3_k10}
\end{table*}

As shown in Tables~\ref{tab:exp1_k10}, \ref{tab:exp2_k10}, and \ref{tab:exp3_k10}, the same trend holds under the larger budget. When both test-time scaling and automatic harness evolution are allowed to run for more iterations, automatic harness evolution still shows no consistent advantage, even on the model with which it performs best, and it continues to generalize poorly on Terminal-Bench.

\section{Experiments on ARC-AGI-3}
\label{sec:appendix_arcagi}
Given the preceding results, we hypothesize that the effectiveness of automatic harness evolution can be more clearly demonstrated on out-of-distribution tasks that remain highly challenging for current agents. We therefore further evaluate on two game benchmarks that present complex, novel simulated environments with iterative evaluation feedback, in which agents must discover how to act and win. Our experimental results show that, in such settings, automatic harness evolution yields clear benefits.

\subsection{Experimental Details}
\label{sec:appendix_arcagi_experimental_details}

\paragraph{Dataset}
ARC-AGI-3 is an interactive benchmark comprising novel, abstract, turn-based environments in which agents must explore, infer goals, build internal models of environment dynamics, and plan effective action sequences, all without explicit instructions. The agent observes a $64 \times 64$ grid in which each cell takes one of 16 colors; a single grid state is referred to as a \textit{frame}. The agent is never told what the objective is and must infer even what constitutes winning. Each game consists of a sequence of levels of increasing difficulty, and levels are scored individually. Humans solve 100\% of the environments, and 25 public demo environments have been released.

ARC-AGI-3 evaluates agents using Relative Human Action Efficiency (RHAE), which measures per-level action efficiency against a human baseline and normalizes it per game. Completing a level is not sufficient on its own: the number of interactions required is what the metric captures. The per-level score is $\min\left(1.15, \left(a_{\text{human}} / a_{\text{agent}}\right)^2\right)$, the squared ratio of human to agent actions, capped slightly above human performance. The per-game score is a level-weighted mean in which later levels receive greater weight. Only environment actions enter this count; internal reasoning and read-only inspection are free under the scoring protocol.

\paragraph{Configuration}
We select the model and reasoning effort so that the tasks are neither saturated nor entirely unsolved, ensuring that differences between methods remain visible. Table~\ref{tab:hparams_arcagi} shows the full configurations.

\begin{table}[h!]
    \centering
    \small
    \setlength{\tabcolsep}{6pt}
    \renewcommand{\arraystretch}{1.05}
    \begin{tabular}{lll}
        \toprule[1pt]
        \textbf{Category} & \textbf{Hyperparameter} & \textbf{Value} \\
        \midrule
        Agent
            & Model                       & Claude Opus 4.8 \\
            & Reasoning effort            & xhigh \\
            & Agent                       & \texttt{claude-code@2.1.220}\\
            \\
            & Model                       & GPT-5.6 Terra \\
            & Reasoning effort            & high \\
            & Agent                       & \texttt{codex@0.145.0} \\
        \midrule
        Implementation
            & Backend                     & Docker \\
            & Max actions                 & 2000 \\
        \bottomrule[1pt]
    \end{tabular}
    \caption{ARC-AGI-3 configurations.}
    \label{tab:hparams_arcagi}
\end{table}

\paragraph{Implementation}
Tables~\ref{tab:arcagi} and~\ref{tab:arcagi_winrate} report two variants of each method. For Sequential Refinement, we evaluate one variant using plain Claude Code CLI or Codex CLI, and one using VISTA*, i.e., VISTA with memory evolution disabled. For Harness Scaling, we evaluate the original VISTA implementation, which evolves memory, and a variant built on VISTA that additionally allows the agent to evolve skills, tools, and hooks.

Each backend is run independently on all 25 public ARC-AGI-3 games. The model receives PNG observations, the current game status and progress, and the set of available actions. Within each backend, the same initial prompt, tools, model, and reasoning setting are used across all games. We detail each variant below.

\begin{itemize}
    \item \textbf{Sequential Refinement} [Claude Code / Codex]: general-purpose multimodal backends, namely Claude Code CLI for Claude Opus 4.8 and Codex CLI for GPT-5.6 Terra. The agent plays and revises its strategy based on the scores it receives, until the budget is exhausted. The harness itself is fixed.
    \item \textbf{Sequential Refinement} [VISTA*]: to isolate harness evolution from test-time scaling, we use a plain version of VISTA with its memory evolution disabled, so that the harness remains unchanged throughout the run. VISTA runs on Claude Code and Codex, adding raw PNG observations, lossless visual memory with inspection tools (\texttt{inspect}, \texttt{read\_pixels}), and context-reset continuation for long sessions. Its memory consists of \texttt{GUIDE.md}, a durable model of the game maintained across levels, and \texttt{WORKING.md}, a scratchpad reset at each level boundary; in VISTA*, both are disabled.
    \item \textbf{Harness Scaling} [VISTA]: the original VISTA implementation, which restores \texttt{GUIDE.md} and \texttt{WORKING.md} together with the tools that maintain them (\texttt{read\_guide}, \texttt{write\_guide}, \texttt{read\_working}, and \texttt{write\_working}). Harness evolution is thus limited to memory.
    \item \textbf{Harness Scaling} [full harness]: built on VISTA, this variant extends evolution beyond memory to the full harness, including skills, tools, and hooks. Concretely, we add \texttt{read\_skill}, \texttt{write\_skill}, \texttt{write\_tool}, \texttt{run\_tool}, and \texttt{write\_hooks}.
\end{itemize}

\paragraph{Win Rate} We present the win rate of all methods on the 25 public games in Table~\ref{tab:arcagi_winrate}. Win rate measures the number of levels the agent completes, regardless of how many actions it takes. The RHAE scores in Table~\ref{tab:arcagi} account for both level completion and the number of actions taken relative to the human baseline.

\begin{table*}[h!]
\centering
\setlength{\tabcolsep}{4pt}
\resizebox{1.0\linewidth}{!}{
\begin{tabular}{lcccc|cccc}
\toprule[1pt]
 & \multicolumn{4}{c|}{\textbf{Claude Opus 4.8}} & \multicolumn{4}{c}{\textbf{GPT-5.6 Terra}}\\
 \cmidrule[0.75pt](lr){2-5} \cmidrule[0.75pt](lr){6-9}
 & \multicolumn{2}{c}{\textbf{Sequential Refinement}} & \multicolumn{2}{c|}{\textbf{Harness Scaling}} & \multicolumn{2}{c}{\textbf{Sequential Refinement}} & \multicolumn{2}{c}{\textbf{Harness Scaling}} \\
  \textbf{Task} & [Claude Code] & [VISTA*] & [VISTA] & [full harness] & [Codex] & [VISTA*] & [VISTA] & [full harness]\\
\midrule
ar25 &5/8 &\underline{6/8}  &\textbf{8/8} &\textbf{8/8} &2/8 &\underline{5/8} &\textbf{8/8} &\textbf{8/8}\\
\addlinespace[1pt]
bp35  &\underline{1/9} &\textbf{3/9} &\textbf{3/9} &\textbf{3/9} &1/9 &\underline{3/9} &\underline{3/9} &\textbf{5/9} \\
\addlinespace[1pt]
cd82  &\underline{2/6} &\textbf{6/6} &\textbf{6/6} &\textbf{6/6} &\underline{0/6} &\textbf{6/6} &\textbf{6/6} &\textbf{6/6}\\
\addlinespace[1pt]
cn04  &\underline{4/6} &\underline{4/6} &\underline{4/6} &\textbf{5/6} &\underline{2/6} &\textbf{4/6} &\textbf{4/6} &\textbf{4/6}\\
\addlinespace[1pt]
dc22  &\underline{2/6} &\textbf{4/6} &\textbf{4/6} &\textbf{4/6} &\underline{1/6} &\textbf{4/6} &\textbf{4/6} &\textbf{4/6}\\
\addlinespace[1pt]
ft09  &\textbf{6/6} &\textbf{6/6} &\textbf{6/6} &\textbf{6/6} &\underline{3/6} &\textbf{6/6} &\textbf{6/6} &\textbf{6/6}\\
\addlinespace[1pt]
g50t  &1/7 &1/7 &\textbf{6/7} &\underline{5/7} &\textbf{2/7} &\underline{1/7} &\textbf{2/7} &\textbf{2/7} \\
\addlinespace[1pt]
ka59  &5/7&\underline{6/7}&\textbf{7/7}&\textbf{7/7} &\underline{1/7} &\textbf{6/7} &\textbf{6/7} &\textbf{6/7}\\
\addlinespace[1pt]
lf52  &2/10&3/10&\underline{4/10}&\textbf{6/10} &2/10 &\underline{3/10} &\textbf{5/10} &\textbf{5/10}\\
\addlinespace[1pt]
lp85  &\underline{3/8} &\textbf{8/8} &\textbf{8/8} &\textbf{8/8} &1/8 &3/8 &\underline{6/8} &\textbf{8/8}\\
\addlinespace[1pt]
ls20  &\underline{4/7}&3/7&\textbf{7/7}&\textbf{7/7} &1/7 &2/7 &\underline{3/7} &\textbf{4/7}\\
\addlinespace[1pt]
m0r0  &\textbf{6/6}&\textbf{6/6}&\textbf{6/6}&\textbf{6/6} &\underline{2/6} &\textbf{6/6} &\underline{2/6} &\textbf{6/6}\\
\addlinespace[1pt]
r11l  &\underline{4/6}&\textbf{6/6}&\textbf{6/6}&\textbf{6/6} &1/6 &\underline{2/6} &\textbf{5/6} &\underline{2/6}\\
\addlinespace[1pt]
re86  &6/8&5/8&\underline{7/8}&\textbf{8/8} &2/8 &\underline{4/8} &\underline{4/8} &\textbf{6/8}\\
\addlinespace[1pt]
s5i5  &\underline{6/8} &2/8 &\underline{6/8} &\textbf{8/8} &1/8 &1/8 &\underline{3/8} &\textbf{6/8} \\
\addlinespace[1pt]
sb26  &\textbf{8/8}&\textbf{8/8}&\textbf{8/8}&\textbf{8/8} &\underline{0/8} &\textbf{8/8} &\textbf{8/8} &\textbf{8/8}\\
\addlinespace[1pt]
sc25  &\textbf{6/6}&\underline{5/6}&\textbf{6/6}&\textbf{6/6} &0/6 &\underline{3/6} &\underline{3/6} &\textbf{4/6}\\
\addlinespace[1pt]
sk48  &\textbf{4/8}&2/8&\textbf{4/8}&\underline{3/8} &\underline{0/8} &\textbf{2/8} &\textbf{2/8} &\underline{0/8}\\
\addlinespace[1pt]
sp80  &1/6&\underline{4/6}&1/6&\textbf{6/6} &0/6 &\underline{1/6} &\textbf{3/6} &\underline{1/6}\\
\addlinespace[1pt]
su15  &3/9&\underline{4/9}&3/9&\textbf{8/9} &3/9 &3/9 &\textbf{9/9} &\underline{7/9}\\
\addlinespace[1pt]
tn36  &0/7&\textbf{3/7}&\underline{1/7}&\underline{1/7} &0/7 &3/7 &\textbf{6/7} &\underline{5/7} \\
\addlinespace[1pt]
tr87  &\underline{5/6}&1/6&\underline{5/6}&\textbf{6/6} &0/6 &\underline{2/6} &1/6 &\textbf{6/6}\\
\addlinespace[1pt]
tu93  &\textbf{9/9}&\textbf{9/9}&\textbf{9/9}&\textbf{9/9} &2/9&5/9&\textbf{9/9}&\underline{8/9}\\
\addlinespace[1pt]
vc33  &\underline{6/7}&4/7&4/7&\textbf{7/7} &0/7&\textbf{4/7}&\textbf{4/7}&\underline{1/7}\\
\addlinespace[1pt]
wa30  &\textbf{7/9}&\underline{3/9}&\textbf{7/9}&\textbf{7/9} &0/9&1/9&\underline{2/9}&\textbf{3/9}\\
\addlinespace[2pt]
\hdashline
\addlinespace[3pt]
\textbf{Average} &4.2/7.3 &4.5/7.3 &\underline{5.4/7.3} &\textbf{6.2/7.3} &1.1/7.3 &3.5/7.3 &\underline{4.6/7.3} &\textbf{4.8/7.3}\\
\bottomrule[1pt]
\end{tabular}
}
\caption{Experimental results on all 25 public ARC-AGI-3 games. We report win rate here; scores are shown in the Table~\ref{tab:arcagi}. The best results are in \textbf{bold} and the second best are \underline{underlined}. [VISTA*] denotes VISTA with memory evolution disabled. [VISTA] adds back memory evolution, as in the original implementation. [full harness] further allows the agent to evolve the entire harness, including skills, tools, and hooks. On ARC-AGI-3, automatic harness evolution yields clear benefits.}
\label{tab:arcagi_winrate}
\end{table*}

\paragraph{Cost} Following VISTA, we cap the budget at 2,000 actions, corresponding to 2,000 calls to \texttt{play}. We also observe that in some cases the agent chooses to exit before exhausting this budget.

\subsection{Which harness artifacts do agents actually use?} \label{sec:appendix_arcagi_artifacts}
To understand where the gains of full-harness evolution come from, we count how often each harness artifact is written and read over the course of a run, averaged across the 25 public games (Table~\ref{tab:arcagi_harness_count}).
Three patterns stand out.

First, tool use dominates every other form of harness interaction \footnote{The nine operations in Table~\ref{tab:arcagi_harness_count} are the fixed interface through which the agent inspects and modifies its harness, not artifacts the agent creates. Counts of \texttt{write\_tool} and \texttt{run\_tool} refer to agent-written tools (e.g.\ \texttt{frame\_diff}, \texttt{match\_symbols}) and exclude calls to the interface operations themselves.}. Agents write roughly 11 tools per game with Claude Opus 4.8 and 20 with GPT-5.6 Terra, and invoke them 132 and 197 times respectively, about ten calls per tool written in both cases. Tool operations alone account for 143 of the 169 harness operations per game with Claude Opus 4.8 and 217 of 260 with GPT-5.6 Terra. This amortization is the mechanism behind the efficiency gains in Section~\ref{sec:arcagi}: the cost of writing a tool is paid once, while the benefit is collected on every subsequent step that would otherwise require manual inspection or search.

Second, giving agents the full harness partially displaces memory. Memory operations fall from 24.2 to 20.0 per game with Claude Opus 4.8 and from 35.4 to 27.0 with GPT-5.6 Terra once skills and tools are available, a reduction of 17\% and 24\% respectively. Rather than accumulating prose descriptions of what they have observed, agents offload the same knowledge into executable artifacts, which can be applied directly instead of reread and reinterpreted.

Third, not all mechanisms are adopted equally. Hooks are almost never written (0.2 and 0.0 per game), suggesting that agents do not spontaneously reach for event-triggered rules even when the interface is available; one of the cases in which hooks do help appears in Appendix~\ref{sec:appendix_arcagi_casestudy}. The two models also differ in how they consume what they write: both write a similar number of skills (4.3 and 3.8 per game), but GPT-5.6 Terra reads them back far more often than Claude Opus 4.8 (12.6 versus 2.0). GPT-5.6 Terra is more active across the board, issuing 260 harness operations per game to Claude Opus 4.8's 169, which suggests differing working styles rather than differing amounts of harness evolution.

We note that these counts measure usage rather than value; a single well-chosen tool may matter more than a dozen marginal ones, and the case studies in Appendix~\ref{sec:appendix_arcagi_casestudy} give a more direct view of which artifacts are load-bearing.

\begin{table*}[h!]
\centering
\resizebox{0.75\linewidth}{!}{
\begin{tabular}{lcc|cc|c}
\toprule[1pt]
 & \multicolumn{2}{c|}{\textbf{Claude Opus 4.8}} & \multicolumn{2}{c|}{\textbf{GPT-5.6 Terra}} & \\
 \cmidrule[0.75pt](lr){2-3} \cmidrule[0.75pt](lr){4-5}
 & \multicolumn{2}{c|}{\textbf{Harness Scaling}} & \multicolumn{2}{c|}{\textbf{Harness Scaling}} & \textbf{Average}\\
  & [VISTA] & [full harness] & [VISTA] & [full harness] &\\
\midrule
\texttt{read\_guide}  &2.7&1.7&10.4&6.4&5.3\\
\addlinespace[1pt]
\texttt{write\_guide}  &6.1&4.4&14.6&9.5&8.7\\
\addlinespace[1pt]
\texttt{read\_working}  &1.0&0.9&1.1&3.0&1.5\\
\addlinespace[1pt]
\texttt{write\_working}  &14.4&13.0&9.3&8.1&11.2\\
\addlinespace[1pt]
\texttt{read\_skill} &-&2.0&-&12.6&7.3\\
\addlinespace[1pt]
\texttt{write\_skill}  &-&4.3&-&3.8&4.1\\
\addlinespace[1pt]
\texttt{write\_tool}  &-&11.4&-&19.6&15.5\\
\addlinespace[1pt]
\texttt{run\_tool}  &-&131.5&-&197.1&164.3\\
\addlinespace[1pt]
\texttt{write\_hooks}  &-&0.2&-&0.0&0.1\\
\addlinespace[2pt]
\hdashline
\addlinespace[3pt]
\textbf{Total} &24.2&169.4&35.4&260.1&122.3\\
\bottomrule[1pt]
\end{tabular}
}
\caption{Average number of harness operations per game on the 25 public ARC-AGI-3 games. Dashes mark operations unavailable under [VISTA], which evolves memory only; the \textbf{Average} column therefore averages over all four columns for memory operations and over the two [full harness] columns otherwise.}
\label{tab:arcagi_harness_count}
\end{table*}

\subsection{Case Study} \label{sec:appendix_arcagi_casestudy}
To illustrate why evolving the full harness helps, we examine several games in which the two Harness Scaling variants diverge most sharply. In each case, the bottleneck turns out to be perceptual rather than strategic, and the ability to write tools is what allows the agent to overcome it.

\begin{figure*}[h!]
\centering
\begin{subfigure}[t]{0.23\linewidth}
    \centering
    \includegraphics[width=\linewidth]{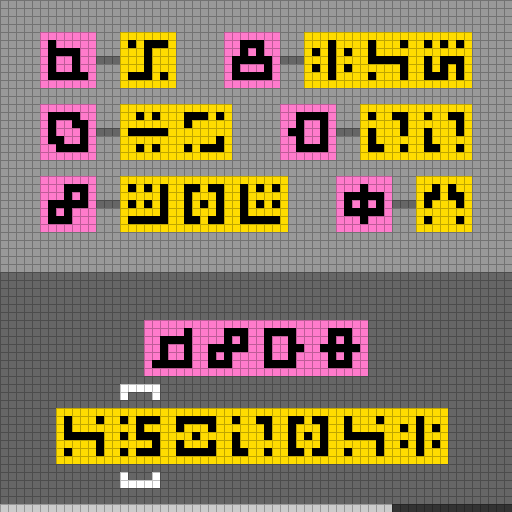}
    \caption{[VISTA], still on level 2 at turn 1{,}415, when the run is terminated.}
    \label{fig:arcagi_tr87_stuck}
\end{subfigure}
\hfill
\begin{subfigure}[t]{0.23\linewidth}
    \centering
    \includegraphics[width=\linewidth]{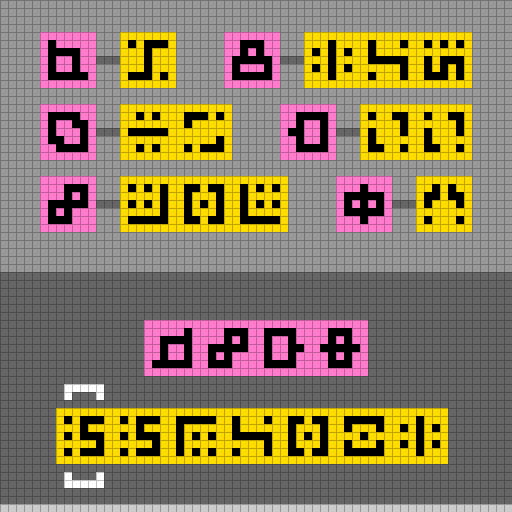}
    \caption{[full harness], entering level 2 at turn 33. The agent writes \texttt{match\_symbols} and \texttt{level2\_status} here.}
    \label{fig:arcagi_tr87_l2start}
\end{subfigure}
\hfill
\begin{subfigure}[t]{0.23\linewidth}
    \centering
    \includegraphics[width=\linewidth]{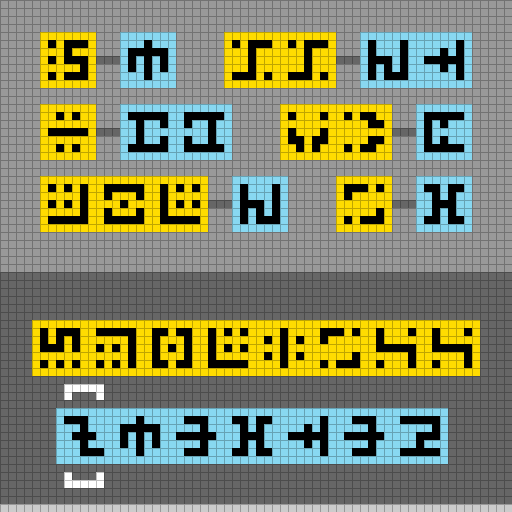}
    \caption{[full harness], level 2 solved at turn 65, only 32 actions later.}
    \label{fig:arcagi_tr87_l2solved}
\end{subfigure}
\hfill
\begin{subfigure}[t]{0.23\linewidth}
    \centering
    \includegraphics[width=\linewidth]{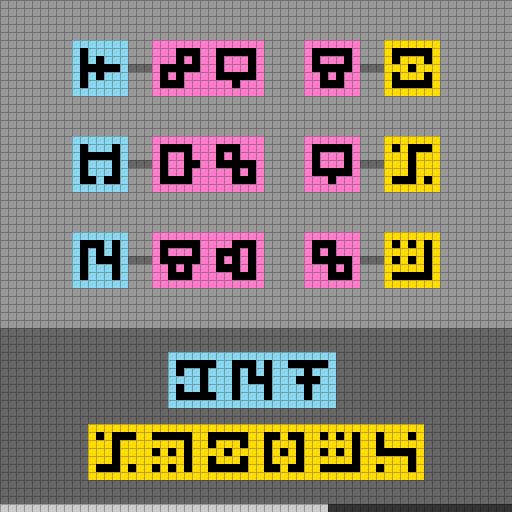}
    \caption{[full harness], game won at turn 256, 6/6 levels.}
    \label{fig:arcagi_tr87_full}
\end{subfigure}
\caption{Progress on \texttt{tr87} with GPT-5.6 Terra. Evolving only memory leaves the agent enumerating wheel positions on level 2 for 1{,}415 actions without solving it (a). Full-harness evolution writes pixel-level perception tools and clears the same level in 32 actions (b--c), finishing the game in 256 actions (d).}
\label{fig:arcagi_tr87}
\end{figure*}

\paragraph{\texttt{tr87} with GPT-5.6 Terra}
Under Harness Scaling [VISTA], which evolves only memory, the agent clears 1 of 6 levels in 1{,}415 actions (Figure~\ref{fig:arcagi_tr87_stuck}). Under Harness Scaling [full harness], it clears all 6 in 256 actions (Figure~\ref{fig:arcagi_tr87}), solving level 2 alone in 32 (Figures~\ref{fig:arcagi_tr87_l2start}--\ref{fig:arcagi_tr87_l2solved}).

The bottleneck is perceptual rather than strategic. The agent misreads the glyphs visually, describing them as six pink $7 \times 7$ input glyphs when they are in fact $5 \times 5$, and, unable to compare glyphs reliably, falls back on a combinatorial search over wheel positions. Its final \texttt{GUIDE.md} records the search rather than the rule:

\begin{quote}
\small
\texttt{Critical empirical search: ALL c5,c7 pairs were rejected for c1..c4=[4,6,3,1] with c6=4 AND with c6=3. Other previously rejected entire c5,c7 families: [5,6,3,1,*,4,*], [6,6,5,5,*,3,*], [6,6,3,1,*,4,*]. Thus direct early matching/segmentation state assumptions are incorrect. Next untested local branch is c6=2, but reconsider prefix/partition.}
\end{quote}

\noindent
This brute-force enumeration over $7$ slots and $7$ states consumes 1{,}320 actions on level 2 alone. Because the harness exposes no way to inspect pixels programmatically, memory evolution can only accumulate prose about a misperception it cannot correct.

The evolution trace (\texttt{private/evolve\_log.jsonl}) shows the agent writing perception tools within the first few turns. At turn 1 it calls \texttt{write\_tool} to create \texttt{components} and \texttt{glyph\_masks}, which extract each card's $5 \times 5$ black-pixel mask exactly, correct on the first attempt. When level 2 begins at turn 32, it writes \texttt{match\_symbols} and \texttt{level2\_status}, canonicalizing each glyph under the eight dihedral transforms and reporting per-slot correctness:

{\small
\begin{verbatim}
    def canon(a):
        z = []; q = a
        for _ in range(4): z.append(q); q = ro(q)
        q = fh(a)
        for _ in range(4): z.append(q); q = ro(q)
        return min(''.join('1' if v else '0' for r in x for v in r) 
        for x in z)
    ...
    return [{'slot': i, 'correct': canon(cur[i]) == canon(t[i]),
    'id': canon(cur[i])} for i in range(7)]
\end{verbatim}
}

\noindent
Over turns 32--64, the agent issues 13 calls to \texttt{level2\_status}, 3 to \texttt{glyph\_masks}, 2 to \texttt{decode\_level}, and 2 to \texttt{analyze\_wheel}, solving level 2 in 32 actions. The insight that the VISTA run never reaches, namely that glyphs match only up to rotation and reflection, is written into the skill \texttt{visual\_substitution.md}: input matches may be rotations or reflections, so compare dihedral orbits.

Across the full run, the agent issues 52 \texttt{write\_tool} calls producing 17 distinct tools, roughly one per level layout, along with 158 \texttt{run\_tool} calls and 2 \texttt{write\_skill} calls. The resulting toolset (\texttt{codex\_visible/tools/index.json}) is: \texttt{analyze\_wheel}, \texttt{classify\_chars}, \texttt{components}, \texttt{decode\_level}, \texttt{glyph\_masks}, \texttt{level2\_status}, \texttt{level3\_parse}, \texttt{level3\_status}, \texttt{level4\_data}, \texttt{level4\_status}, \texttt{level5\_data}, \texttt{level5\_packed}, \texttt{level5\_probe}, \texttt{level6\_parse}, \texttt{match\_symbols}, and \texttt{orbit\_ids}.

The binding constraint on this game is reading $5 \times 5$ glyphs from a $64 \times 64$ frame, which a twenty-line pixel tool eliminates outright. When evolution is restricted to memory, the agent can only write prose about what it has already misperceived; when it can write tools, it repairs the perceptual channel itself and the strategic problem becomes tractable.

\paragraph{\texttt{lp85} with Claude Opus 4.8}
On this game, both Harness Scaling variants clear all eight levels, so the difference lies entirely in efficiency: memory evolution requires 271 actions, while full-harness evolution requires 104. The human baseline is 388 actions in total.

\begin{table}[h]
\centering
\small
\begin{tabular}{lcccccccc c}
\toprule
& L1 & L2 & L3 & L4 & L5 & L6 & L7 & L8 & Total \\
\midrule
Human baseline & 17 & 38 & 31 & 16 & 41 & 60 & 26 & 159 & 388 \\
Harness Scaling [VISTA] & 8 & 11 & 19 & 22 & 49 & 23 & 6 & 133 & 271 \\
Harness Scaling [full harness] & 8 & 9 & 19 & 13 & 10 & 20 & 5 & 20 & 104 \\
\bottomrule
\end{tabular}
\caption{Per-level action counts on \texttt{lp85} with Claude Opus 4.8. Both variants win all eight levels; the divergence is concentrated in level 8.}
\label{tab:lp85_actions}
\end{table}

As Table~\ref{tab:lp85_actions} shows, the two runs track each other closely through level 7 and diverge sharply on level 8: 133 actions under memory evolution against 20 under full-harness evolution, with a human baseline of 159.

\texttt{lp85} is a permutation puzzle. Arrow controls rotate rings and rows of tiles, and every special tile must be brought to rest on its matching sparkle cursor. Solving a level therefore amounts to finding a short word in the group generated by the available rotations, which is a search problem the agent can express exactly once it can read the board reliably.

\begin{figure*}[h!]
\centering
\begin{subfigure}[t]{0.3\linewidth}
    \centering
    \includegraphics[width=\linewidth]{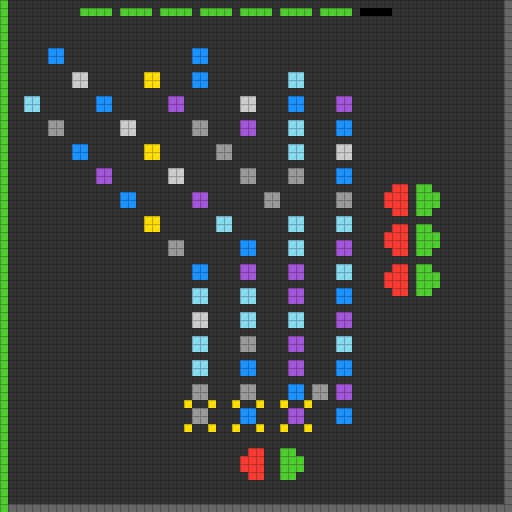}
    \caption{[VISTA], level 8 at turn 139. Reasoning about the rotations in prose, the agent needs 133 further actions to finish.}
    \label{fig:lp85_default_l8}
\end{subfigure}
\hspace{0.02\linewidth}
\begin{subfigure}[t]{0.3\linewidth}
    \centering
    \includegraphics[width=\linewidth]{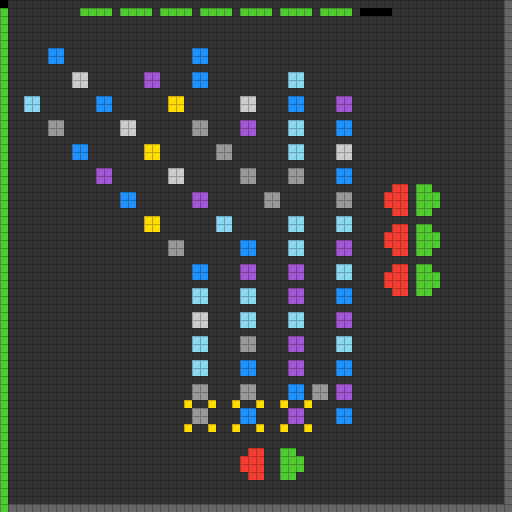}
    \caption{[full harness], level 8 at turn 85. The BFS solver \texttt{perm3} has just been written for this layout.}
    \label{fig:lp85_evolve_l8}
\end{subfigure}
\hspace{0.02\linewidth}
\begin{subfigure}[t]{0.3\linewidth}
    \centering
    \includegraphics[width=\linewidth]{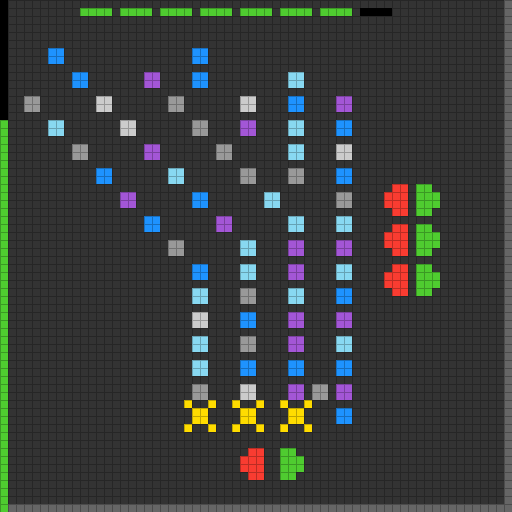}
    \caption{[full harness], solved at turn 104, only 20 actions later and well inside the human baseline of 159.}
    \label{fig:lp85_evolve_win}
\end{subfigure}
\caption{Level 8 of \texttt{lp85} with Claude Opus 4.8. Both variants win the game, but evolving only memory (a) leaves the agent enumerating wheel positions by hand, while a purpose-written solver (b--c) reduces the same level from 133 actions to 20.}
\label{fig:arcagi_lp85}
\end{figure*}

Rather than reasoning about rotations in prose, the agent incrementally constructs a parser and a search procedure over the first ninety turns:

\begin{center}
\small
\begin{tabular}{ll}
\toprule
Turn & Asset written \\
\midrule
$\sim$0  & \texttt{write\_tool ring} (3 revisions; parses ring membership from the grid) \\
$\sim$4  & \texttt{write\_tool dump} \\
$\sim$8  & \texttt{write\_tool parse2} (3 revisions) \\
$\sim$10 & \texttt{write\_tool solve2} (first BFS solver over permutation states) \\
$\sim$17 & \texttt{write\_tool blocks} \\
$\sim$19 & \texttt{write\_tool solve3} \\
$\sim$29 & \texttt{write\_skill loop\_puzzle} (1{,}712 characters) \\
$\sim$36 & \texttt{write\_tool blocks} \\
$\sim$38 & \texttt{write\_tool solve4} \\
$\sim$51 & \texttt{write\_tool solve5} \\
$\sim$60 & \texttt{write\_tool cycle} \\
$\sim$81 & \texttt{write\_tool solve7} \\
$\sim$88 & \texttt{write\_tool perm3} (level 8 solver; 20 actions) \\
$\sim$91 & \texttt{write\_skill loop\_puzzle} (3{,}759 characters) \\
\bottomrule
\end{tabular}
\end{center}

Each \texttt{solveN} is a breadth-first search over the group generated by the arrow controls:

{\small
\begin{verbatim}
    def rot(order, d):
        p = list(range(N)); n = len(order)
        for k in range(n): p[order[k]] = order[(k + d) % n]
        return p
    
    moves = {"V_up":    rot(V, 1),  "V_down": rot(V, -1),
             "H_right": rot(H, 1),  "H_left": rot(H, -1)}
    # BFS from start to goal over (pos_a, pos_b) states; returns the
    move list
\end{verbatim}
}

Over the full run the agent issues 16 \texttt{write\_tool} calls, 64 \texttt{run\_tool} calls, and 2 \texttt{write\_skill} calls. The accompanying skill, \texttt{loop\_puzzle.md}, records the control semantics and the visual conventions the tools depend on:

\begin{quote}
\small
\texttt{Only ACTION6 (click) is used. Arrows are the controls. GREEN arrow = rotate +1 (CW / right / down / outward). RED arrow = $-1$. \dots Special tiles: yellow = grid11, orange = grid12. Sparkle cursor = 4 single yellow/orange marks at a tile's outer corners $(r-1,c-1)$, $(r-1,c+2)$, $(r+2,c-1)$, $(r+2,c+2)$.}
\end{quote}

The effect is not specific to Claude Opus 4.8. On the same game with GPT-5.6 Terra, full-harness evolution wins all eight levels in 872 actions (per level: 12, 107, 60, 47, 33, 280, 40, 293), whereas memory evolution wins only six, exhausting the 2{,}000-action budget after spending 1{,}413 actions stuck on level 6 (per level: 10, 163, 195, 108, 68, 1{,}413, 43, 0).

Unlike \texttt{tr87}, where tool evolution unlocks levels the agent could not otherwise reach, here it converts an expensive manual search into a cheap programmatic one.

\paragraph{\texttt{ls20} with Claude Opus 4.8}
On \texttt{ls20}, full-harness evolution with Claude Opus 4.8 is the only configuration in our experiments to clear \texttt{ls20} in full, winning all seven levels in 666 actions against a human baseline of 776.

\texttt{ls20} is a block-pushing and key-delivery maze that layers several mechanics on top of one another: fog that hides unexplored regions, entities that move independently of the agent, and portal launchers that slide the player across the board. Levels 5 and 6 introduce the moving entities, and this is precisely where every other run stalls.

\begin{figure*}[h!]
\centering
\begin{subfigure}[t]{0.23\linewidth}
    \centering
    \includegraphics[width=\linewidth]{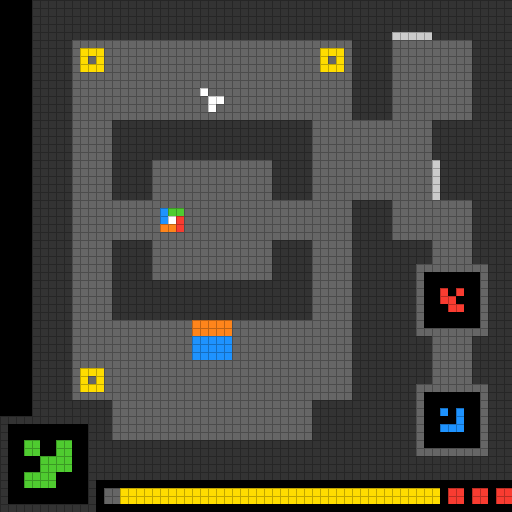}
    \caption{Level 5, turn 378. \texttt{route\_gen} and \texttt{intercept} have just been written to plan around portal launchers and the orbiting entity.}
    \label{fig:ls20_l5}
\end{subfigure}
\hfill
\begin{subfigure}[t]{0.23\linewidth}
    \centering
    \includegraphics[width=\linewidth]{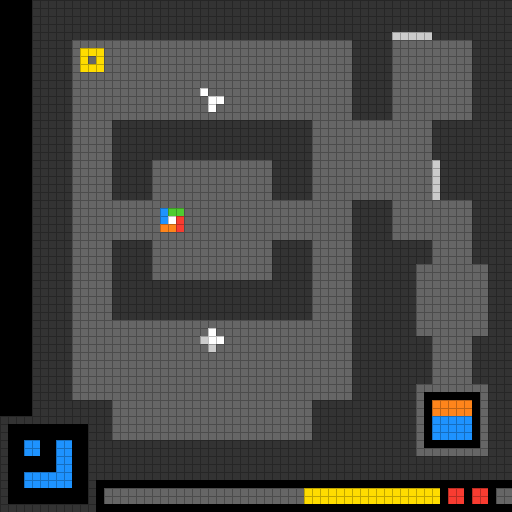}
    \caption{Level 6 cleared, turn 547. The multi-entity planner \texttt{multi} handles the level where every other run stalls.}
    \label{fig:ls20_l6}
\end{subfigure}
\hfill
\begin{subfigure}[t]{0.23\linewidth}
    \centering
    \includegraphics[width=\linewidth]{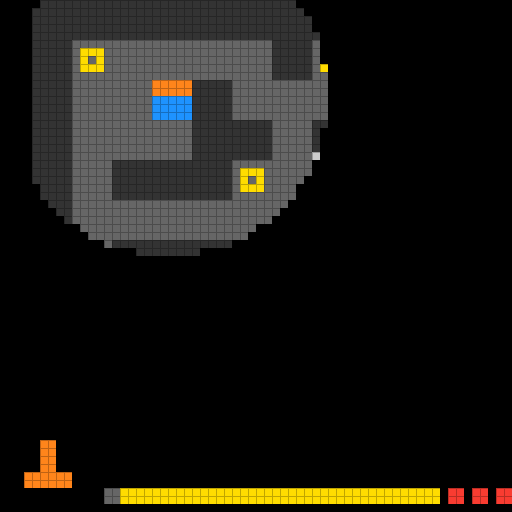}
    \caption{Level 7, turn 548. The fog maze, which no other configuration in our experiments reaches.}
    \label{fig:ls20_l7}
\end{subfigure}
\hfill
\begin{subfigure}[t]{0.23\linewidth}
    \centering
    \includegraphics[width=\linewidth]{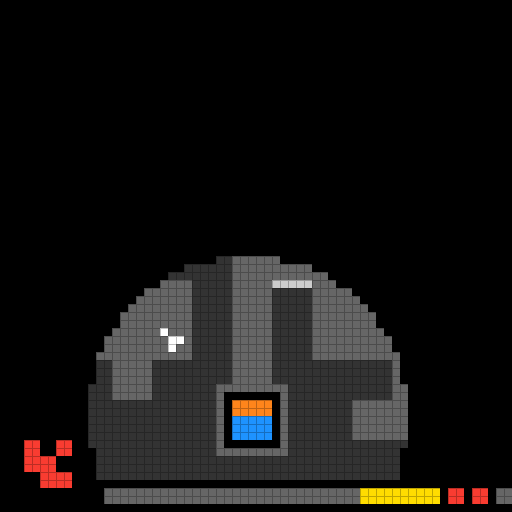}
    \caption{Game won at turn 666, 7/7 levels against a human baseline of 776 actions.}
    \label{fig:ls20_win}
\end{subfigure}
\caption{Progression of full-harness evolution on \texttt{ls20} with Claude Opus 4.8, the only configuration in our experiments to clear the game. Planning tools written around turn 380 carry the agent through the moving-entity levels (a--b) and into the fog maze (c), which it completes in 119 actions (d).}
\label{fig:arcagi_ls20}
\end{figure*}

Unlike the previous two cases, this run exercises tools, skills, and hooks together. The trace shows perception tools appearing first, followed by planners that handle progressively richer dynamics:

\begin{center}
\small
\begin{tabular}{ll}
\toprule
Turn & Asset written \\
\midrule
$\sim$0   & \texttt{write\_tool ascii}, \texttt{colors}, \texttt{map} (render grid as walkable/wall map) \\
$\sim$19  & \texttt{write\_tool tiles}, \texttt{path} (BFS over the $5\times5$-tile grid) \\
$\sim$19  & \texttt{write\_hooks} (rule \texttt{blocked\_move}) \\
$\sim$73  & \texttt{write\_tool path2} (BFS with avoid-list) \\
$\sim$167 & \texttt{write\_tool shape} \\
$\sim$350 & \texttt{write\_tool raw}, \texttt{route5} (launch-aware BFS for level 5) \\
$\sim$376 & \texttt{write\_tool route\_gen} (auto-detects portal launchers from bar edges) \\
$\sim$385 & \texttt{write\_tool intercept} (time-expanded BFS to collide with an orbiting entity) \\
$\sim$393 & \texttt{write\_tool multi} (multi-entity, turn-synchronized planner) \\
$\sim$547 & \texttt{write\_skill block-and-delivery} (1{,}424 characters) \\
$\sim$553 & \texttt{write\_tool scan} \\
$\sim$561 & \texttt{write\_tool tbfs} (classifies fog vs.\ wall; lists reachable tiles) \\
$\sim$570 & \texttt{write\_tool launch} (simulates the portal slide tile by tile) \\
$\sim$665 & \texttt{write\_skill block-and-delivery} (2{,}766 characters) \\
\bottomrule
\end{tabular}
\end{center}

The hook written at turn 19 converts a recurring failure into an automatic diagnostic, firing three times over the run (at steps 201, 454, and 485):

{\small
\begin{verbatim}
    [{
      "name": "blocked_move",
      "event": "after_play",
      "when": {"action_in": ["ACTION1","ACTION2","ACTION3","ACTION4"],
               "frame_unchanged_steps_at_least": 1},
      "note": "Frame unchanged: block likely hit a wall (did not move).
               Re-check tile map/position.",
      "block": false
    }]
\end{verbatim}
}

In total the agent writes 17 tools, issues 287 \texttt{run\_tool} calls, and writes 2 skills and 1 hook. The accumulated skill records the visual conventions the planners rely on:
\begin{quote}
\small
\texttt{The movable BLOCK renders as a solid $5\times5$ tile and may be TWO-TONE (e.g.\ 2 rows orange over 3 rows blue). This two-tone is NORMAL and persists as it moves; it is NOT an animation glitch. The PANEL (bottom-left box) is the AUTHORITATIVE current key: read its SHAPE and COLOR with the \texttt{shape} tool per color \dots\ LOCK = framed box with a colored shape inside \dots\ There can be MULTIPLE locks; ALL must be filled. A lock only accepts a key matching EXACTLY (shape AND color).}
\end{quote}

The tools written at turns 376--393 (\texttt{route\_gen}, \texttt{intercept}, and \texttt{multi}) are not perception aids but planning procedures: they reason about portal dynamics and about entities that move while the agent does. This is the capability that the runs stalling on levels 5 and 6 lack, and it is not one that editing a memory file can supply. Taken with the previous two cases, the pattern is that full-harness evolution helps along three distinct axes, repairing perception on \texttt{tr87}, replacing manual search with programmatic search on \texttt{lp85}, and adding planning machinery here.

\section{EdgeBench}
\label{sec:appendix_edgebench}

\subsection{Experimental Details}
\label{sec:appendix_edgebench_experimental_details}

\paragraph{Dataset} The \textit{Interactive Games \& Simulators} category contains real games designed for human players, where proficient humans typically invest tens of hours to master the mechanics. The state spaces are enormous and each run is procedurally distinct, so agents face strong out-of-distribution pressure. Agents must develop and refine strategies through high-frequency interaction across many episodes. We evaluate on the five tasks in which the agent writes and iteratively refines a game-playing program (a Lua, Squirrel, or Python bot) that is scored by running it in the game simulator.

The agent can submit its current artifacts to the judge at any time, subject to a cooldown, and use the returned score and feedback to guide further work. The EdgeBench harness also scores the workspace from the host side at fixed intervals. These snapshots go through the same hidden judge and are recorded for analysis, but their results are never shown to the agent. This lets EdgeBench measure improvement between explicit submissions while keeping agent-visible feedback separate from evaluator-only measurement. The final score is the best-so-far maximum across all judge-scored evaluations from either source, on a 0--100 scale.

\begin{itemize}
    \item{\textbf{dcss\_dungeon\_ai}}: A classic roguelike dungeon crawler. The task is to write a Lua bot that plays a fixed character (a Minotaur Berserker wielding a mace). Scoring runs 5 hidden seeds with a 900-second wall-clock limit per game, averaging DCSS's built-in score (experience, depth, items, kills). There is no turn limit, so score roughly equals turns-per-second times points-per-turn: the bot must use batch commands (auto-explore \texttt{o}, auto-descend \texttt{G>}) to play fast while staying alive at deeper levels.
    \item{\textbf{nethack\_dungeon\_agent}}: The 1987 roguelike classic, played through NLE (NetHack Learning Environment). The task is to write \texttt{agent.py} exporting \texttt{get\_action(obs, info, step)}, which returns an integer action each step given a $21 \times 79$ character/glyph grid plus a status bar.
    \item{\textbf{openrct2\_theme\_park\_ai}}: A theme-park management task (open-source RollerCoaster Tycoon 2). The agent writes an ES5 JavaScript plugin (\texttt{my\_plugin.js}) that uses \texttt{context.executeAction} to build rides, open them, set prices, and hire staff across small/medium/large scenarios. Anti-cheat rules forbid modifying guest generation or directly mutating game state; violations score 0.
    \item{\textbf{openttd\_transport\_ai}}: A transport-business management task (open-source Transport Tycoon Deluxe). The agent writes an AI player in Squirrel under the NoAI framework (\texttt{main.nut}) that builds roads, railways, airports, and routes, buys vehicles, and manages loans, as the only company on the map. Scoring runs 20 in-game years on 5 seed maps and averages company value (vehicles at $1.5\times$ purchase price $+$ infrastructure $+$ cash $-$ loans). Because scores are averaged, the worst seed drags the result down the most (e.g., one bankrupt seed), so robustness matters more than peak performance on a single seed.
    \item{\textbf{wesnoth\_tactical\_ai}}: A turn-based tactics game. The agent writes a Lua AI plus WML configuration (\texttt{my\_custom\_ai.cfg} defining candidate actions and aspects such as aggression, caution, and village value) to play side 1 against the default AI. The score is $80 \times \text{win rate} + 20 \times \text{average efficiency}$ (efficiency $=$ baseline turns / actual turns, counted only for wins), over 3 maps $\times$ 10 seeds $\times$ 2 repeats $=$ 60 games. A set of APIs is banned (\texttt{wml\_actions}, \texttt{io.open}, etc.); violations score 0 and are flagged as hacks.
\end{itemize}

\paragraph{Configuration}
Following the official EdgeBench evaluation protocol~\citep{zhu2026edgebench}, we evaluate Claude Opus 4.8 and GPT-5.4 and explore whether automatic harness evolution can be beneficial on its game tasks. Table~\ref{tab:hparams_edgebench} lists the full configuration.

\begin{table}[h!]
    \centering
    \small
    \setlength{\tabcolsep}{6pt}
    \renewcommand{\arraystretch}{1.05}
    \begin{tabular}{lll}
        \toprule[1pt]
        \textbf{Category} & \textbf{Hyperparameter} & \textbf{Value} \\
        \midrule
        Agent
            & Model                       & Claude Opus 4.8 \\
            & Reasoning effort            & high \\
            & Agent                       & \texttt{claude-code@2.1.159}\\
            & Context window              & 1{,}000{,}000 tokens\\
            \\
            & Model                       & GPT-5.4 \\
            & Reasoning effort            & xhigh \\
            & Agent                       & \texttt{codex@0.130.0} \\
        \midrule
        Backend
            & Backend                     & Kubernetes (k8s) \\
            & Timeout                     & 43{,}200\,s \\
            & Evaluation interval         & 1{,}800\,s \\
            & Submission cooldown         & 120\,s \\
        \midrule
        Resources
            & Work CPU limit              & 4 \\
            & Work memory limit           & 16\,GB \\
            & Judge CPU limit             & 4 \\
            & Judge memory limit          & 8\,GB \\
        \midrule
        Trigger
            & Submission Step        & 3 \\
            & Time Interval                & 7{,}200\,s \\
        \bottomrule[1pt]
    \end{tabular}
    \caption{EdgeBench configurations.}
    \label{tab:hparams_edgebench}
\end{table}

\paragraph{Implementation}
Under Harness Scaling, we implement a \textit{periodic} harness update mechanism: the meta agent is triggered to update the harness after every $N$ agent submissions or $T$ minutes, whichever comes first. This design accommodates tasks of different natures: on some tasks a single exploration takes a long time, while on others the agent makes frequent, quick submissions. We set $N = 3$ and $T = 120$ so that the harness is revisited at reasonable intervals throughout the 12-hour window.

Each update proceeds as follows. When triggered, the meta agent pauses the live agent. A summarizer condenses the agent's trajectory, and the meta agent edits the harness based on this summary, with access to the raw trajectory when more detail is needed. If the meta agent makes a change and the change passes validation, it writes a \texttt{sync\_request}, the new harness is installed, and the live agent resumes with it. Otherwise, the live agent resumes with the unchanged harness.

\paragraph{Cost}
We match the budget at 12 hours of wall-clock time for both Sequential Refinement and Harness Scaling. Under Sequential Refinement, the full 12 hours is spent by the agent running and revising its solution in response to feedback. Under Harness Scaling, the same 12 hours covers both harness editing and gameplay: the time spent on trajectory summarization and harness editing counts against the budget, and the agent remains idle while an update is in progress. Both settings therefore consume the same wall-clock budget.

\begin{figure*}[h!]
\centering
\includegraphics[width=1.0\linewidth]{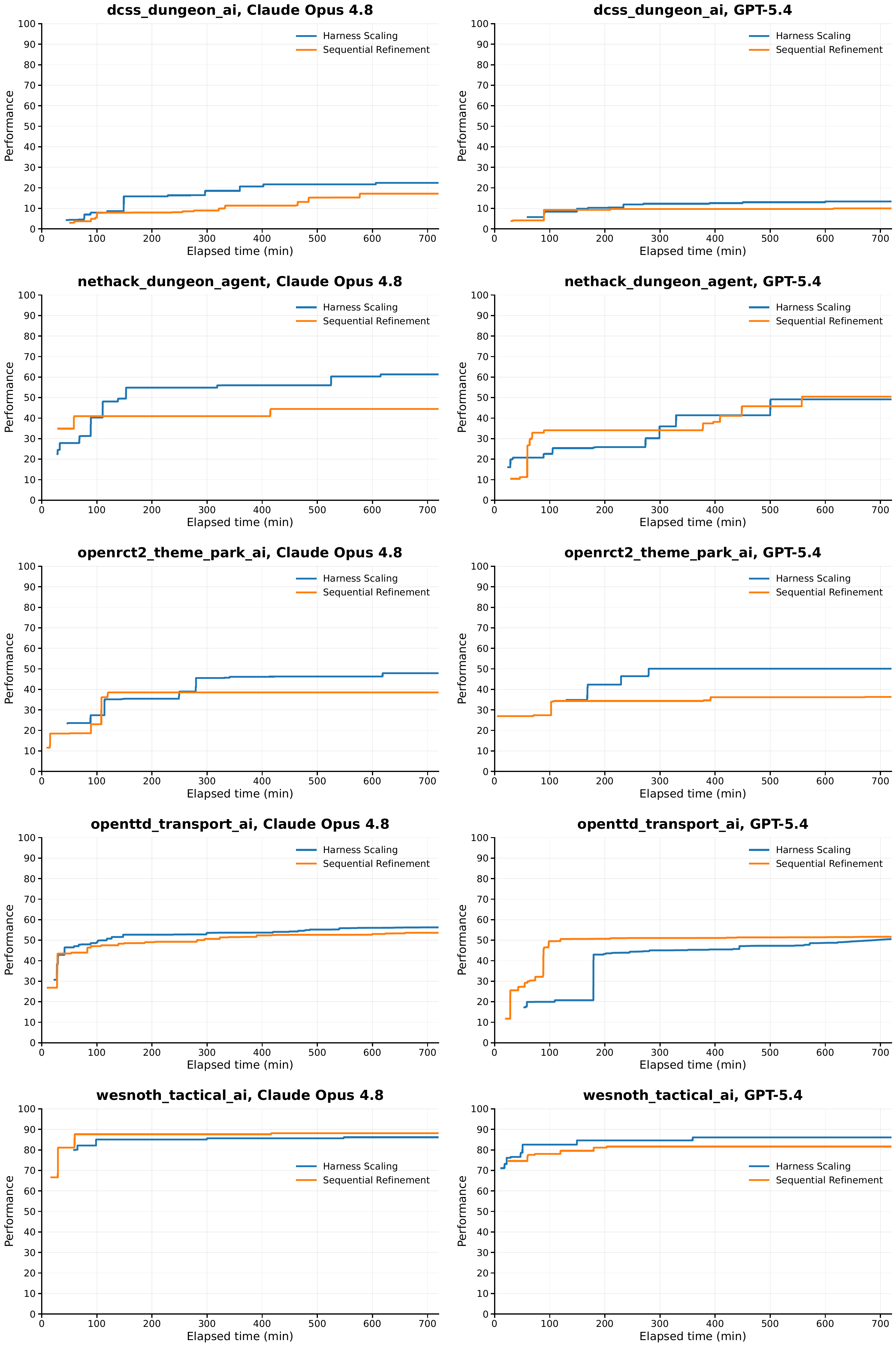}
\caption{EdgeBench performance over the 12-hour window.}
\label{fig:edgebench_fig}
\end{figure*}

\paragraph{Performance over the full 12-hour window}
Table~\ref{tab:edgebench-12hr} and Figure~\ref{fig:edgebench_fig} report performance over the full 12-hour window. Each entry gives the best score achieved up to that point in the run, so the reported curves are monotonically non-decreasing in time. It is important to compare not only final performance but also how quickly each method reaches a given score, which matters when the budget is the binding constraint.

\begin{table}[h!]
\centering
\begin{tabular}{lccccccc}
\toprule[1pt]
 & 2h & 4h & 6h & 8h & 10h & 12h & \textbf{Best} \\
\midrule
\textbf{dcss\_dungeon\_ai} [Claude Opus 4.8] &&&&&&& \\
\hspace{2em} Sequential Refinement & 7.9 & 8.0 & 11.3 & 15.2 & 17.1 & 17.1 & \textbf{17.1} \\
\hspace{2em} Harness Scaling & 8.7 & 16.3 & 18.4 & 21.6 & 22.4 & 22.4 & \textbf{22.4} \\
\textbf{dcss\_dungeon\_ai} [GPT-5.4] &&&&&&& \\
\hspace{2em} Sequential Refinement & 9.3 & 9.6 & 9.6 & 9.6 & 9.6 & 9.9 & \textbf{9.9} \\
\hspace{2em} Harness Scaling & 8.3 & 11.8 & 12.1 & 12.9 & 13.2 & 13.3 & \textbf{13.3} \\
\addlinespace[2pt]
\hdashline
\addlinespace[2pt]
\textbf{nethack\_dungeon\_agent} [Claude Opus 4.8] &&&&&&& \\
\hspace{2em} Sequential Refinement & 40.8 & 40.8 & 40.8 & 44.3 & 44.3 & 44.3 & \textbf{44.3} \\
\hspace{2em} Harness Scaling & 48.0 & 54.7 & 55.9 & 55.9 & 60.2 & 61.3 & \textbf{61.3} \\
\textbf{nethack\_dungeon\_agent} [GPT-5.4] &&&&&&& \\
\hspace{2em} Sequential Refinement  & 34.0 & 34.0 & 37.3 & 45.7 & 50.4 & 50.4 & \textbf{50.4} \\
\hspace{2em} Harness Scaling & 25.3 & 25.8 & 41.3 & 41.3 & 49.0 & 49.0 & \textbf{49.0} \\
\addlinespace[2pt]
\hdashline
\addlinespace[2pt]
\textbf{openrct2\_theme\_park\_ai} [Claude Opus 4.8] &&&&&&& \\
\hspace{2em} Sequential Refinement & 38.4 & 38.4 & 38.4 & 38.4 & 38.4 & 38.4 & \textbf{38.4} \\
\hspace{2em} Harness Scaling & 35.0 & 35.4 & 46.0 & 46.2 & 46.2 & 47.7 & \textbf{47.8} \\
\textbf{openrct2\_theme\_park\_ai} [GPT-5.4] &&&&&&& \\
\hspace{2em} Sequential Refinement & 34.2 & 34.2 & 34.2 & 36.1 & 36.1 & 36.2 & \textbf{36.2} \\
\hspace{2em} Harness Scaling  & 27.5 & 46.3 & 50.0 & 50.0 & 50.0 & 50.0 & \textbf{50.0} \\
\addlinespace[2pt]
\hdashline
\addlinespace[2pt]
\textbf{openttd\_transport\_ai} [Claude Opus 4.8] &&&&&&& \\
\hspace{2em} Sequential Refinement & 47.4 & 49.1 & 51.5 & 52.5 & 52.9 & 53.5 & \textbf{53.5} \\
\hspace{2em} Harness Scaling & 50.7 & 52.7 & 53.5 & 54.8 & 55.9 & 56.2 & \textbf{56.2} \\
\textbf{openttd\_transport\_ai} [GPT-5.4] &&&&&&& \\
\hspace{2em} Sequential Refinement & 49.5 & 50.9 & 51.0 & 51.4 & 51.4 & 51.6 & \textbf{51.6} \\
\hspace{2em} Harness Scaling & 20.6 & 43.8 & 45.2 & 47.2 & 48.6 & 50.5 & \textbf{50.5} \\
\addlinespace[2pt]
\hdashline
\addlinespace[2pt]
\textbf{wesnoth\_tactical\_ai} [Claude Opus 4.8] &&&&&&& \\
\hspace{2em} Sequential Refinement & 87.5 & 87.5 & 87.5 & 88.0 & 88.0 & 88.0 & \textbf{88.0} \\
\hspace{2em} Harness Scaling & 85.0 & 85.0 & 85.5 & 85.5 & 86.0 & 86.0 & \textbf{86.0} \\
\textbf{wesnoth\_tactical\_ai} [GPT-5.4] &&&&&&& \\
\hspace{2em} Sequential Refinement & 78.0 & 81.5 & 81.5 & 81.5 & 81.5 & 81.5 & \textbf{81.5} \\
\hspace{2em} Harness Scaling & 82.5 & 84.5 & 84.5 & 86.0 & 86.0 & 86.0 & \textbf{86.0} \\
\bottomrule[1pt]
\end{tabular}
\caption{EdgeBench performance over the 12-hour window.}
\label{tab:edgebench-12hr}
\end{table}

\end{document}